%% file: main.tex
\documentclass[10pt,twocolumn,letterpaper]{article}

\usepackage{cvpr}              
\usepackage{graphicx}
\usepackage{svg}
\usepackage{booktabs}
\usepackage{colortbl}
\usepackage{multirow}
\usepackage{multicol}
\usepackage{tabularx}
\usepackage{makecell}
\usepackage{diagbox}
\usepackage{wrapfig}
\usepackage{threeparttable} 
\usepackage{tablefootnote}
\usepackage{float}
\usepackage{balance}
\usepackage{algorithm}
\usepackage{algpseudocode}

\input{preamble}
\definecolor{cvprblue}{rgb}{0.21,0.49,0.74}
\usepackage[pagebackref,breaklinks,colorlinks,allcolors=cvprblue]{hyperref}

\def\paperID{} 
\def\confName{CVPR}
\def\confYear{2026}

\title{TAP: A Token-Adaptive Predictor Framework for Training-Free Diffusion Acceleration}

\author{Haowei Zhu\textsuperscript{\rm 1,2}, Tingxuan Huang\textsuperscript{\rm 1}, Xing Wang\textsuperscript{\rm 2}, Tianyu Zhao\textsuperscript{\rm 2}, Jiexi Wang\textsuperscript{\rm 2}, Weifeng Chen\textsuperscript{\rm 2}, \\Xurui Peng\textsuperscript{\rm 2}, Fangmin Chen\textsuperscript{\rm 2}, Junhai Yong\textsuperscript{\rm 1}, Bin Wang\textsuperscript{\rm 1}\thanks{Corresponding author: {wangbins@tsinghua.edu.cn}}\\
\textsuperscript{\rm 1} Tsinghua University
\textsuperscript{\rm 2} ByteDance}

\begin{document}
\maketitle
\input{sec/0_abstract}    
\input{sec/1_intro}
\input{sec/2_related_works}
\input{sec/3_method}
\input{sec/4_experiment}

\input{sec/5_conclusion}

{
    \small
    \bibliographystyle{ieeenat_fullname}
    \bibliography{main}
}

\input{sec/X_suppl}


\end{document}

%% file: sec/0_abstract.tex
\begin{abstract}
Diffusion models achieve strong generative performance but remain slow at inference due to the need for repeated full-model denoising passes. We present \textbf{Token-Adaptive Predictor} (TAP), a training-free, probe-driven framework that adaptively selects a predictor for each token at every sampling step. TAP uses a single full evaluation of the model's first layer as a low-cost probe to compute proxy losses for a compact family of candidate predictors (instantiated primarily with Taylor expansions of varying order and horizon), then assigns each token the predictor with the smallest proxy error. This per-token “probe-then-select’’ strategy exploits heterogeneous temporal dynamics, requires no additional training, and is compatible with various predictor designs. TAP incurs negligible overhead while enabling large speedups with little or no perceptual quality loss. Extensive experiments across multiple diffusion architectures and generation tasks show that TAP substantially improves the accuracy–efficiency frontier compared to fixed global predictors and caching-only baselines.

\end{abstract}

%% file: sec/1_intro.tex
\section{Introduction}
\label{sec:intro}

Diffusion Models (DMs)~\cite{DM} have become a cornerstone of generative modeling, delivering state-of-the-art results in image synthesis~\cite{StableDiffusion} and video generation~\cite{blattmann2023SVD}. Advances in scalable architectures such as Diffusion Transformers (DiT)~\cite{DiT} have further improved fidelity, but the sequential nature of diffusion sampling remains a fundamental bottleneck: every denoising step requires a full forward pass through an often large and computationally expensive model. This per-step full-model requirement creates a stark trade-off between generation quality and inference efficiency that grows more severe as models and datasets scale. Existing reductions in sampling steps (e.g., DDIM~\cite{songDDIM}) mitigate some temporal dependency but cannot fully avoid trajectory truncation errors. Complementary approaches aim to reduce per-step cost by reusing or approximating intermediate features across timesteps, yet they face their own limitations.

Many recent caching and prediction-based accelerators exploit temporal redundancy in features across adjacent timesteps~\cite{selvaraju2024fora,zou2024accelerating,chen2024delta-dit}. TaylorSeer~\cite{liuReusingForecastingAccelerating2025} formalizes temporal forecasting with Taylor-style expansions to predict multi-step feature evolution and achieves substantial savings at moderate acceleration ratios. However, prior methods typically apply a single, \emph{global} prediction policy to all tokens and timesteps, overlooking an important empirical fact: the appropriate approximation complexity varies across tokens and over time. Low-order Taylor approximations are often sufficient for tokens that evolve slowly, but tokens that experience large displacements or rapid dynamics typically require higher-order or alternative predictors to avoid significant truncation (or extrapolation) errors. Applying a single global predictor throughout the sampling process for every sample therefore risks error accumulation and severe quality degradation at aggressive acceleration ratios. Furthermore, adaptively selecting the appropriate predictor for each token remains an open and nontrivial challenge.

To address these limitations, we propose \textit{Token-Adaptive Predictor} (TAP), a token-adaptive diffusion acceleration framework that dynamically selects, for every token at each sampling step, the most suitable predictor from a compact family of candidates. TAP is founded on a simple, low-overhead insight: input perturbation to the model and the resulting output error are highly correlated, so a single full evaluation of the model's first layer can serve as an informative probe of a token's susceptibility to prediction error. Using this probe, TAP computes proxy losses for multiple candidate predictors in parallel and assigns each token the predictor with the smallest proxy error. This ``probe-then-select'' strategy resembles model-ensemble~\cite{sagi2018ensemble} thinking in that it leverages the complementary strengths of different predictors to assign the best predictor per token. Note that it is agnostic to the internal design of the candidate predictors and therefore can integrate a heterogeneous set of methods.

In this work we instantiate the candidate family primarily with Taylor predictors obtained by varying expansion order and prediction horizon (distance from the expansion point). Different tokens often achieve their best predictions at different orders or in different local neighborhoods of the current timestep, so this design provides a simple yet expressive predictor pool. The per-token probe-then-select paradigm delivers three key advantages. First, it exploits token-level heterogeneity in temporal dynamics to allocate the most appropriate predictor to each token, improving approximation fidelity with negligible extra computation cost. Second, unlike prior adaptive schemes~\cite{liu2024timestep, liu2025speca} that require manually tuned thresholds, TAP makes decisions based on the \textit{relative} proxy errors between predictors and thus requires no extra hand-designed thresholds. Third, the framework is architecturally flexible: the candidate set can include not only classical Taylor expansions of varying order and horizon, but also other prediction-style methods (e.g., FoCa~\cite{zhengFoCa2025}, FreqCa~\cite{liu2025freqca}) or numerically stable polynomial regressors. Importantly, the added prediction overhead is small, predictors reduce to simple pointwise operations on cached feature tensors, and multiple predictors only require a single first-layer evaluation to validate in parallel. Specifically, compared to FLUX.1-dev with 50 steps, TAP introduces only an additional 0.1 \,GB of GPU memory (approximately 0.3\% of the original model) while achieving an acceleration of 6.24$\times$ without loss in perceptual quality.

We evaluate TAP across multiple diffusion architectures and generation tasks and show that per-token adaptive prediction substantially improves the accuracy–efficiency frontier relative to fixed global predictors and naive caching baselines. In summary, our contributions are:
\begin{itemize}
  \item \textbf{Token-adaptive prediction framework.} We introduce TAP, a probe-driven per-token predictor selection framework that adaptively assigns the optimal predictor from a set of candidates. We show that a single full evaluation of the model's first layer provides an effective proxy for predictor selection.
  \item \textbf{Taylor predictor family.} We identify that superior predictions arise at different Taylor orders and prediction horizons, and propose a family of Taylor-based predictors that together cover a diverse range of token dynamics.
  \item \textbf{Comprehensive evaluation.} Extensive experiments across diffusion models and tasks demonstrate that TAP consistently improves the accuracy–efficiency trade-off compared to fixed predictors and existing baselines.
\end{itemize}

%% file: sec/2_related_works.tex
\section{Related Works}
\label{sec:works}
Diffusion models~\cite{sohl2015deep,ho2020DDPM} have achieved remarkable success in high-fidelity image and video generation, but their sequential denoising process imposes substantial computational costs. Efforts to accelerate diffusion inference broadly fall into two categories: reducing the number of sampling timesteps, and reducing per-step computation via model-level or feature-level optimizations. Below we review representative works in these directions and situate our method within this landscape.

\subsection{Sampling Timestep Reduction}
A major research direction aims to reduce the number of denoising iterations while preserving sample quality. DDIM~\cite{songDDIM} introduced deterministic samplers to compress sampling trajectories, and the DPM-Solver family~\cite{lu2022dpm,lu2022dpm++,zheng2023dpmsolvervF} leverages higher-order ODE solvers to decrease the required steps with controlled integration error. While these methods are effective at lowering step counts, they typically trade off fidelity when pushed to extreme speedups and do not directly address the high per-step cost of modern, large denoising networks.
Orthogonal to solver-based acceleration, knowledge distillation techniques~\cite{salimans2022progressive,meng2022on,yin2024improved} compress the sampling process into far fewer network evaluations. These approaches often recover performance comparable to the original model but require additional fine-tuning. In contrast, our method is training-free and can be applied on top of both vanilla and distilled samplers. We show that it produces substantial wall-clock speedups without degrading generation quality.

\subsection{Denoising Network Acceleration}
Reducing per-step cost can be achieved through either model compression or feature reuse and prediction.

\paragraph{Model Compression.}
Compression methods, including pruning~\cite{structural_pruning_diffusion,zhu2024dipgo}, quantization~\cite{10377259,shang2023post,kim2025ditto}, distillation~\cite{li2024snapfusion}, and token-reduction techniques~\cite{bolya2023tomesd,kim2024tofu}, reduce the computational footprint of denoising networks. These methods often require retraining or finetuning and balance compute savings against expressive capacity and fidelity loss.

\noindent\textbf{Feature Caching and Prediction.}
Feature caching leverages temporal redundancy across timesteps to avoid recomputing stable activations. Methods such as DeepCache~\cite{ma2024deepcache}, $\Delta$-DiT~\cite{chen2024delta-dit}, TeaCache~\cite{liu2024timestep}, and ToCa~\cite{zou2024accelerating} cache and reuse intermediate features, while DiTFastAttn~\cite{yuan2024ditfastattn} reduces intra-step redundancy. However, these ``cache-then-reuse'' approaches degrade when timestep gaps widen or feature dynamics shift. Forecasting-based methods predict future features to mitigate this issue: TaylorSeer~\cite{liuReusingForecastingAccelerating2025} applies Taylor-style expansions for multi-step prediction, FreqCa~\cite{liu2025freqca} introduces frequency-aware caching with Hermite interpolation for high-frequency components, and SpeCa~\cite{liu2025speca} adopts a draft-then-verify strategy based on stepwise error estimation. However, these methods either lack token-level adaptivity or rely on hand-crafted thresholds, which limit scalability and robustness.

\begin{figure*}[!ht]
    \centering
    \includegraphics[width=\linewidth]{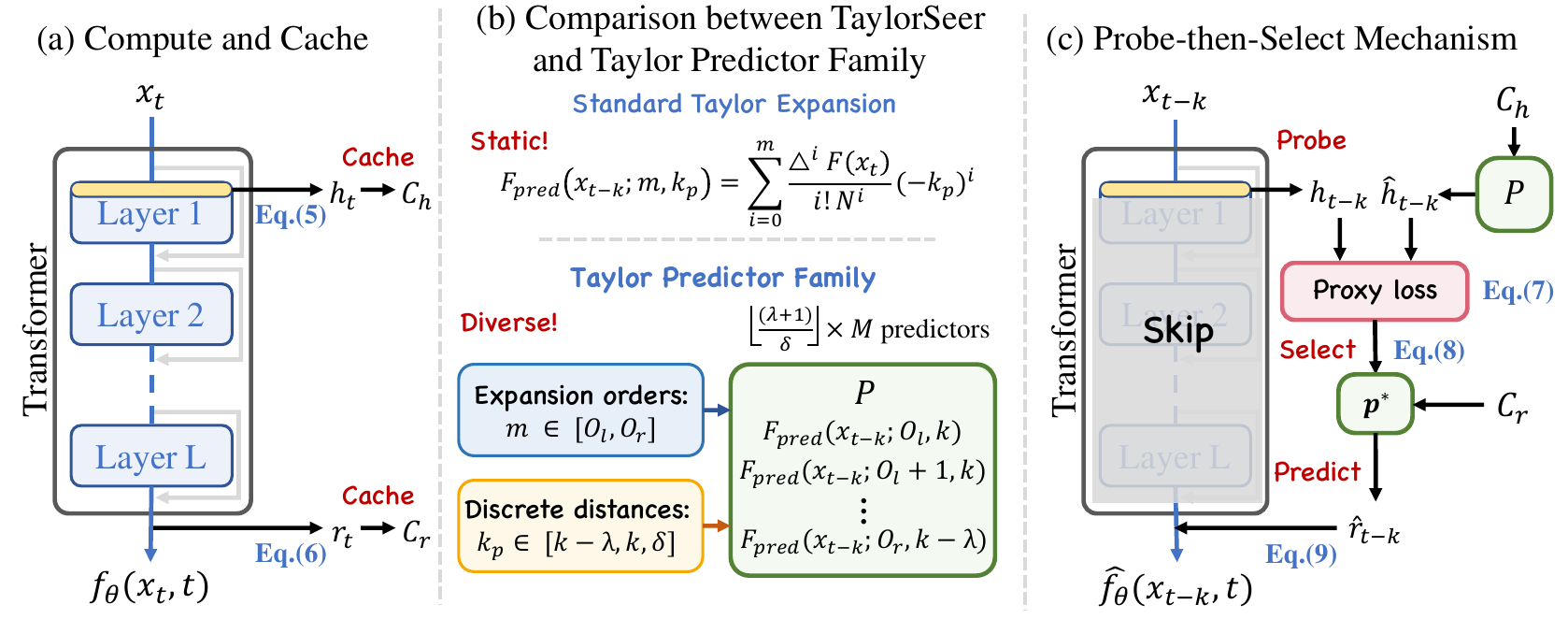}
    \caption{Overview of the TAP framework. (a) \textbf{Compute and cache}: at the first step of each $N$-step window TAP performs a full evaluation and caches the input--output residual and the first-layer modulated input for later probing. (b) \textbf{Taylor predictor family}: we construct a compact set of predictors by varying Taylor expansion order and prediction horizon to cover diverse token dynamics. (c) \textbf{Probe-then-select}: the cached modulated input is used as a lightweight probe to score candidate predictors per token; the selected predictor's output replaces the full model computation for that token and step.}
    \label{fig:method}
\end{figure*}

\noindent\textbf{Our Approach.}
TAP is inspired by forecasting-based caching and speculative execution but departs from prior work in three key ways: 

\noindent (1) \textbf{Token-adaptive prediction:} rather than applying a static, global predictor to all tokens and timesteps, TAP dynamically selects predictors on a per-token basis to capture local temporal heterogeneity. 

\noindent(2) \textbf{Probe-driven scoring:} a single full evaluation of the model's first layer serves as a lightweight probe to estimate proxy losses and score candidate predictors in parallel; unlike prior methods that use shallow-layer errors only to decide whether to execute or skip downstream computation~\cite{liu2024timestep,bu2025dicache}, TAP uses the shallow-layer probe to \emph{select} the best predictor for each token during skipped steps.

\noindent (3) \textbf{Threshold-free design:} decisions are based on relative proxy errors across candidates, eliminating the need for manually tuned thresholds. Together, these properties let TAP mitigate error accumulation, improve per-token prediction fidelity, and complement existing caching, forecasting, and speculative acceleration techniques.

%% file: sec/3_method.tex
\section{Method}\label{sec:method}

\subsection{Preliminary}
\paragraph{Diffusion Models.} 
Diffusion models learn a reverse denoising process that maps Gaussian noise back to data by inverting a forward noising process. The forward step adds Gaussian noise according to a schedule \(\{\alpha_t\}\):
\begin{equation}
\mathbf{x}_t = \sqrt{\alpha_t}\,\mathbf{x}_0 + \sqrt{1-\alpha_t}\,\epsilon_t,\quad \epsilon_t\sim\mathcal{N}(0,I),
\end{equation}
Thus after \(T\) steps the sample is essentially noise. The learned reverse step denoises \(\mathbf{x}_t\) via a neural model \(f_\theta(\mathbf{x}_t,t)\), a commonly used discrete update is:
\begin{equation}
\mathbf{x}_{t-1}=\frac{1}{\sqrt{\alpha_t}}\Big(\mathbf{x}_t - \frac{1-\alpha_t}{\sqrt{1-\bar\alpha_t}}\;f_\theta(\mathbf{x}_t,t)\Big) + \sigma_t\epsilon,
\end{equation}
with \(\bar\alpha_t=\prod_{s=1}^t\alpha_s\) and \(\sigma_t\) controlling injected noise. Equivalently, sampling can be interpreted via reverse SDEs or ODE solvers. Because sampling involves many sequential denoising steps, each requiring a full computation of a large model, inference cost becomes a major bottleneck.

\paragraph{Feature Caching in Diffusion Models.} 
Feature caching reduces per-step cost by avoiding redundant computation across timesteps. Two main strategies have been explored:

\textbf{Reuse-based caching} directly reuses stored activations from a previous timestep \(t\):
\begin{equation}
\mathcal{F}(\mathbf{x}_{t-k})\gets \mathcal{F}(\mathbf{x}_t), \qquad \forall k\in[1, N-1],
\end{equation}
which can skip intermediate computation and yield up to \(N\!\times\) speedup. However, reuse assumes temporal stability. As the timestep gap \(k\) grows, approximation error accumulates rapidly and output quality degrades.

\textbf{Forecast-based caching} aims to predict future features rather than simply copying past ones. We formulate the forecast as a function that accepts the expansion order \(m\) and the temporal offset \(k\) as inputs:
\begin{equation}
\mathcal{F}_{\text{pred}}(\mathbf{x}_{t-k}; m, k_p)
\;=\;\sum_{i=0}^{m}\frac{\Delta^i \mathcal{F}(\mathbf{x}_t)}{i!\,N^i}(-k_p)^i,
\label{eq:taylor}
\end{equation}
where \(\Delta^i\mathcal{F}\) denotes the \(i\)-th finite difference (with \(\Delta^0\mathcal{F}=\mathcal{F}\) by convention), \(N\) is the temporal normalization factor and $k_p$ denotes the prediction distance ($k_p = k$ in~\citep{liuTaylorSeer2025}).
Forecasting is more accurate than direct reuse for modest \(k\), but it relies on local smoothness and well-behaved higher-order differences. In practice, truncation and numerical instability cause prediction error to grow quickly with \(k\), especially under noisy or complex dynamics.

Moreover, both reuse and global-forecast approaches typically apply the same strategy uniformly across all tokens. This ignores a key empirical fact: \emph{tokens differ in temporal sensitivity}. Some tokens (e.g., smooth background regions) remain stable and are well-served by low-cost reuse or low-order predictors, while others (e.g., edges, moving objects) require higher-order or alternative predictors. Long-range mistakes on a small set of sensitive tokens can disproportionately degrade overall generation quality. These observations motivate token-adaptive, cost-aware prediction schemes that select prediction complexity per token and per step.

\subsection{Overview of TAP}
To overcome the limitations of prior diffusion accelerators that apply a fixed global predictor at every sampling step, we propose the \textbf{Token-Adaptive Predictor} (TAP), a probe-driven framework that makes \emph{per-token} predictor choices. As illustrated in Figure~\ref{fig:method}, TAP dynamically assigns the predictor with the smallest proxy error to each token at every step, reducing overall error by exploiting token-level heterogeneity in temporal dynamics and validating candidates with a lightweight first-layer probe. Below we summarize TAP's core components: how we compute and cache compact proxies, the construction of a diversified Taylor predictor family, and the probe-then-select mechanism used to pick predictors per token and step.

Let $\mathbf{r}_t\in\mathbb{R}^{B\times N_x\times D}$ denote cached features at timestep $t$ for a batch of $B$ samples, each containing $N_x$ tokens of dimension $D$. Our goal is to predict features at future timesteps $t-k$, where $k\in\{1,\dots,N - 1\}$, using previously cached full evaluations while avoiding expensive full forward passes for every token and step.

\paragraph{Compute and Cache.}
At the beginning of every $N$-step denoising window, TAP performs one full model evaluation and accelerates the following $N-1$ steps through feature prediction. During this full step, TAP caches the modulated first-layer input and the global residual between the model’s input and output, which are later used to estimate the feature evolution. Specifically, given the input $\textbf{x}_t$ at timestep $t$, the modulated first-layer input and residual are defined as:
\begin{equation}
\mathbf{h}_t = \mathrm{Modulate}\big(\mathrm{Norm}_1(\mathbf{x}_t),\, \mathbf{s}_t,\, \mathbf{g}_t\big),
\label{eq:modulated_input}
\end{equation}
\begin{equation}
\mathbf{r}_t = f_\theta(\mathbf{x}_t, t) - \mathbf{x}_t.
\label{eq:residual}
\end{equation}
where $\textbf{s}_t$ and $\textbf{g}_t$ denote the learned shift and scale parameters at timestep $t$, $\mathrm{Norm}_1(\cdot)$ is the first-layer normalization, and $f_{\theta}$ represents the denoising network. These cached tensors serve as efficient proxies for subsequent probe-based scoring and prediction without requiring full downstream propagation.

TAP then leverages these cached features to perform token-level forecasting using a compact set of candidate predictors, selecting the optimal one for each token based on probe-computed proxy losses. This mechanism enables high-accuracy, training-free acceleration while maintaining generation fidelity.

\paragraph{Taylor Predictor Family.} 
During the reverse denoising process of diffusion models, hidden features evolve smoothly along the sampling trajectory, which enables local prediction based on previous features, as shown in Eq.(\ref{eq:taylor}). However, a single predictor cannot fit all tokens equally well. Recent studies such as FreqCa~\cite{liu2025freqca} highlight that low-frequency and high-frequency tokens exhibit distinct temporal dynamics, thus requiring different predictors. Besides, ToCa~\cite{zou2024accelerating} selectively reuses only tokens with small important scores. 

To capture this token-level heterogeneity, we construct a \textbf{predictor family} with varying levels of complexity. Specifically, we design $M$ Taylor expansion predictors of different orders by adjusting $m \in [O_l, O_r]$ in Eq.(\ref{eq:taylor}). Low-order predictors are more robust when feature evolution is abrupt or discontinuous, while high-order predictors better approximate smooth and stable dynamics. 

Moreover, different tokens may exhibit different convergence ranges for Taylor expansions. If the prediction horizon $k_p$ in Eq.(\ref{eq:taylor}) exceeds a token's convergence radius, Taylor-based predictions, especially high-order ones, can diverge and amplify accumulated error, leading to unstable forecasts.
To mitigate this, we further diversify predictors by varying the prediction distance $k_p$ from $k-\lambda$ to $k$ (where $k$ is the number of steps since the last full computation), and discretize this range with an interval $\delta$. 

In total, our Taylor predictor family contains $ \left\lfloor \frac{(\lambda+1)}{\delta} \right\rfloor
 \times M$ candidate predictors, where $\lambda$ and $\delta$ control the discretized prediction-horizon range and step, and $M$ control the available Taylor orders (we include orders $0, 1,2$). By default we use $M=3$, $\lambda=4$, and $\delta=1$, yielding $\lfloor(4+1)/1\rfloor\times3=15$ predictors. In practice TAP is not sensitive to these hyperparameters: a compact, diversified pool suffices for strong, consistent gains across models and benchmarks, as shown empirically in Section~\ref{sec:experiments}.

\paragraph{Probe-then-Select Mechanism.}
After constructing a set of candidate predictors, an open challenge is how to robustly and efficiently combine them at inference time. Recent work shows that shallow-layer activations can serve as an effective cache indicator~\cite{liu2024timestep,cheng2025paraattention}. Following this insight, we use the modulated first-layer input $\mathbf{h}_t$ (Eq. (\ref{eq:modulated_input})) as a proxy for predictor quality: a predictor whose \emph{predicted} modulated input is closer to the \emph{actual} modulated input is likely to yield a more accurate downstream prediction.

Concretely, let $\mathcal{P}$ denote the predictor family and let $\widehat{\mathbf{h}}_{t,p}\in\mathbb{R}^{B\times N_x\times D}$ be the modulated-input predicted by predictor $p\in\mathcal{P}$ for the current step at time $t$. We compute a lightweight, per-token proxy error between each predictor's prediction and the actual activated modulated input. For token index $(b,n)$ the proxy loss for predictor $p$ is
\begin{equation}
\mathcal{L}^{b,n}_{p} \;=\; d\big(\widehat{\mathbf{h}}_{t,p}^{b,n},\,\mathbf{h}_{t}^{b,n}\big),
\label{eq:proxy_error_general}
\end{equation}
where \(d(\cdot,\cdot)\) denotes a chosen distance metric (e.g., L1, cosine, or MSE). In this work we empirically use cosine distance. Given the proxy losses, we select the best predictor for each token as follows:
\begin{equation}
p^{\star,b,n} \;=\; \arg\min_{p\in\mathcal{P}} \; \mathcal{L}^{b,n}_p.
\label{eq:token_selection}
\end{equation}
All predictions $\widehat{\mathbf{h}}_{t,p}$ are produced in parallel from cached full-step features, and the losses in Eq.(\ref{eq:proxy_error_general}) are computed elementwise over tokens and batch entries, making the probe-and-select
stage highly efficient. Once $p^{\star,b,n}$ is determined, TAP
applies the corresponding predictor for that token in the remainder of the network for the current step. Denoting the per-token residual predicted by predictor $p^{\star,b,n}$ as $\widehat{\mathbf{r}}_{t, p^{\star,b,n}}^{b,n}$, the predicted per-token residuals assemble into the full predicted residual 
$\widehat{\mathbf{r}}_t$, and the model's final predicted output for the current step is:
\begin{equation}
\widehat{f}_{\theta}(\mathbf{x}_t, t) \;=\; \mathbf{x}_t + \widehat{\mathbf{r}}_t.
\label{eq:predict}
\end{equation}
A pseudo-algorithm of TAP is provided in Algorithm~\ref{alg:tap_concise}.
Two practical benefits of our method are as follows. First, because selection uses only \emph{relative} proxy errors across predictors (Eq.(\ref{eq:token_selection})), TAP does not require any hand-tuned thresholds or calibration heuristics. Second, although predictors differ, they all derive from the same distribution of previously computed full-step features. Replacing a token's downstream computation with a prediction from that shared distribution is compatible with the intrinsic overridability of diffusion sampling~\cite{levin2025differential}, which permits local substitutions in intermediate states without disrupting global inference. Together, these properties yield a scalable, threshold-free mechanism for per-token predictor selection that mitigates error accumulation while remaining training-free.

\begin{algorithm}[t]
\caption{Pseudocode of TAP}
\label{alg:tap_concise}
\begin{algorithmic}[1]
\Require denoiser $f_\theta$, predictor set $\mathcal{P}$, window $N$, distance $d(\cdot,\cdot)$
\State $C_h \leftarrow \varnothing,\; C_r \leftarrow \varnothing$ \Comment{\textcolor{gray}{compact cache: first-layer modulated input and residual}}
\For{$t \leftarrow T$ \textbf{downto} $1$}
  \State $\mathbf{x}_t\leftarrow$ current model input
  \State $\mathbf{h}^{t}\leftarrow \mathrm{Modulate}(\mathrm{Norm}_1(\mathbf{x}_t),\mathbf{s}_t,\mathbf{g}_t)$ \Comment{\textcolor{gray}{Eq.(\ref{eq:modulated_input})}}
  \If{$t \bmod N = 0$}
    \State $\mathbf{r}_{t}\leftarrow f_{\theta}(\mathbf{x}_t,t)-\mathbf{x}_t$ \Comment{\textcolor{gray}{full residual (Eq.(\ref{eq:residual}))}}
    \State $C_h\leftarrow \mathbf{h}_{t},\; C_r\leftarrow \mathbf{r}_{t}$ \Comment{\textcolor{gray}{store compact proxies}}
    \State use $ f_{\theta}(\mathbf{x}_{t},t)$ as the model output for this step
  \Else

  \ForAll{$p\in\mathcal{P}$} \Comment{\textcolor{gray}{parallel prediction from cached proxies (e.g., Taylor variants)}}
   \State $\widehat{\mathbf{h}}_{t,p} \leftarrow \mathrm{Predict}(p, C_h)$ \Comment{\textcolor{gray}{(Eq.(\ref{eq:taylor}))}}
  \EndFor

  \ForAll{tokens $(b,n)$}
    \State $p^{\star,b,n}\leftarrow \arg\min_{p} \; d\big(\widehat{\mathbf{h}}_{t,p}^{b,n},\,\textbf{h}_{t}^{b,n}\big)$ \Comment{\textcolor{gray}{proxy error (Eq.(\ref{eq:proxy_error_general})) and selection (Eq.(\ref{eq:token_selection}))}}
    \State $\widehat{\mathbf{r}}_{t}^{b,n} \leftarrow \mathrm{Predict}(p^{\star,b,n}, C_r)$ \Comment{\textcolor{gray}{(Eq.(\ref{eq:taylor}))}}
  \EndFor
  \State $\widehat{\mathbf{r}}_{t} \leftarrow \widehat{\mathbf{r}}_{t}^{b,n}$ \Comment{\textcolor{gray}{assemble per-token residuals}}
  \State $\widehat f_{\theta}(\mathbf{x}_{t},t)\leftarrow \mathbf{x}_{t} + \widehat{\mathbf{r}}_{t}$ \Comment{\textcolor{gray}{derive the predicted output (Eq.(\ref{eq:predict}))}}
  \State use $\widehat f_{\theta}(\mathbf{x}_{t},t)$ as the model output for this step
  \EndIf
\EndFor
\end{algorithmic}
\end{algorithm}

\subsection{Complexity Analysis}
Standard diffusion sampling requires $T$ sequential steps each performing a full forward pass. TAP reduces this cost by introducing token-adaptive predictions and a lightweight probe-based verification instead of executing full downstream computation at every step. The computation introduced by TAP consists of one full evaluation per $N$-step window, parallel prediction of a compact candidate set (implemented with low-cost pointwise and small-polynomial operations), and the first-layer activation used as the probe. These prediction and verification operations are inexpensive in practice: on FLUX.1-dev with 50 steps TAP adds only 0.1\,GB of GPU memory (0.3\% of the original model) and incurs roughly 0.015\% additional FLOPs compared to a single global-predictor baseline. Importantly, TAP stores only the per-step residual and the modulated first-layer input, so the extra memory is constant with respect to model depth (i.e., $O(1)$), whereas methods that cache every layer incur $O(L)$ memory overhead~\cite{liuTaylorSeer2025, zou2024accelerating}. Wall-clock latency gains depend on model size and hardware, we report measured latency results in the experiments.

%% file: sec/4_experiment.tex
\section{Experiment}
\label{sec:experiments}

\subsection{Experimental Settings}
\paragraph{Model Configurations.}
We evaluate TAP on multiple state-of-the-art diffusion architectures to demonstrate broad applicability and robustness. Specifically, we test on \textbf{FLUX.1-dev}~\cite{flux2024}, \textbf{Qwen-Image}~\cite{wu2025qwenimagetechnicalreport} (and their few-step distilled variants), and a large-scale video generation model, \textbf{HunyuanVideo}~\cite{kong2024hunyuanvideo}. For controlled comparisons we reimplement or adopt public checkpoints of representative acceleration baselines, including FORA~\cite{selvaraju2024fora}, TeaCache  (CVPR'25)~\cite{liu2024timestep}, SpeCa (ACM MM'25)~\cite{liu2025speca}, and TaylorSeer (ICCV'25)~\cite{liuReusingForecastingAccelerating2025}. 
\paragraph{Model Configurations.} For text-to-image evaluation we use DrawBench~\cite{saharia2022drawbench} and report semantic alignment and perceptual quality via ImageReward~\cite{xu2023imagereward} and CLIP Score~\cite{hessel2021clipscore}; for reconstruction-style metrics we additionally report PSNR, SSIM~\cite{wang2004imagequality} and LPIPS~\cite{zhangUnreasonableEffectivenessDeep2018}. Video experiments follow the VBench~\cite{huang2024vbench} protocol: for each of 950 prompts we generate 5 videos under different seeds (4,750 videos total) at 480p (2s duration, 17 frames at 8 FPS). Latency and FLOPs are measured on the corresponding benchmarks using NVIDIA H20 GPUs (single node). We report the average wall‑clock latency and FLOPs per generated sample with a batch size of 1.

\subsection{Image Generation}

\subsubsection{FLUX.1-dev}
\input{tables/1_flux}

We evaluate TAP on FLUX.1-dev~\cite{flux2024} across multiple acceleration regimes and report generation quality, perceptual metrics, average latency, and FLOPs (Table~\ref{table:FLUX-Metrics}). Direct-reuse methods (FORA, TeaCache) exhibit noticeable quality degradation, and global-forecast methods (TaylorSeer, SpeCa) also degrade at long cache horizons (e.g., $N=7$). By contrast, TAP attains a superior accuracy–efficiency frontier relative to these direct-reuse and global-forecast baselines: for instance, at $6.24\times$ acceleration, TAP preserves perceptual and fidelity metrics with no measurable quality loss while yielding large latency and FLOPs reductions. This improvement stems from TAP’s token-adaptive assignment of predictors, which preserves sample quality at matched computational budgets. On the distilled model FLUX-Schnell, we further validate the effectiveness of TAP, achieving up to $2\times$ acceleration without noticeable quality loss. Surprisingly, TAP sometimes yields slightly higher ImageReward and CLIP (e.g., 0.99 and 31.19 at $6.24\times$ vs.\ 0.97 and 30.86 at $4.16\times$). This may stem from TAP's token-adaptive prediction benefiting from larger prediction horizons, which for some prompts produces cleaner or more consistent outputs.

\subsubsection{Qwen-Image}

\input{tables/4_qwen_image}
On Qwen-Image~\cite{wu2025qwenimagetechnicalreport} (Table~\ref{table:qwen-image-Metrics}), TAP yields consistent improvements in both quality preservation and latency reduction compared to TaylorSeer and other cache-based methods. At a $3.57\times$ speedup, TAP attains an ImageReward of 1.23 (vs.\ 1.18 for TaylorSeer). On perceptual metrics TAP also shows clear gains: at $3.57\times$ acceleration TAP's PSNR exceeds TeaCache by about $1.1$\,dB and TaylorSeer by about $2.1$\,dB. Moreover, on the distilled Qwen-Image-Lightning variant TAP achieves substantial speedups with only minor quality degradation. These results indicate that TAP better preserves perceptual quality at high speedups while delivering meaningful latency reductions.

\subsection{Video Generation}

\input{tables/7_hunyuan}

\begin{figure}[h]
\centering
\includegraphics[width=0.9\linewidth]{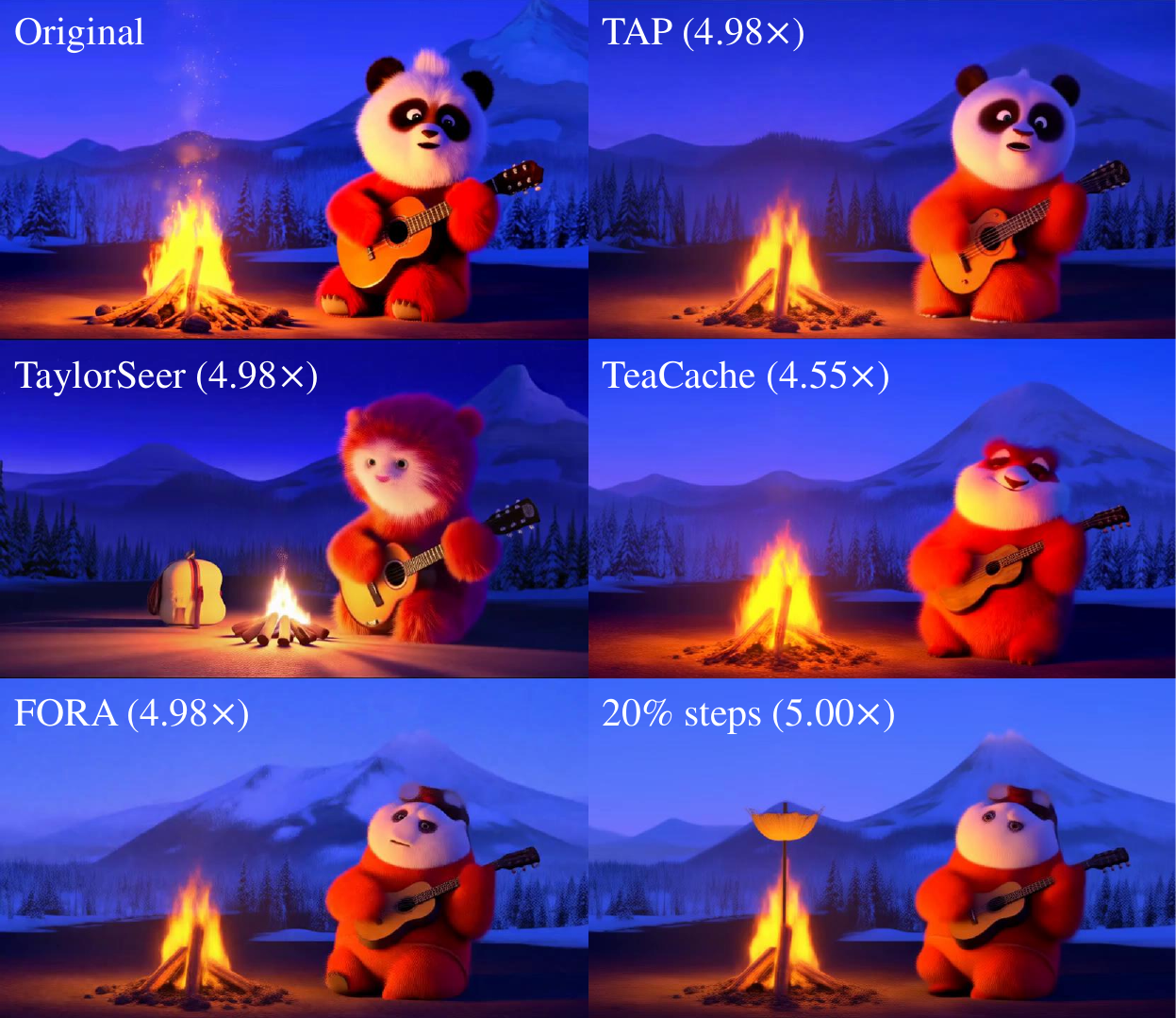}
\caption{Visualization of video generation. \textit{``A happy fuzzy \textcolor{red}{panda} playing guitar nearby a campfire, snow mountain in the background".}}
\label{fig:video_vis}
\vspace{-2mm}
\end{figure}

Table~\ref{table:HunyuanVideo-Metrics} demonstrates that TAP attains the best accuracy–efficiency trade-off on HunyuanVideo~\cite{kong2024hunyuanvideo}. With a cache window of $N=6$, TAP achieves a $4.98\times$ speedup and the highest VBench score (65.46), corresponding to only a $ 1.7\%$ drop versus the unaccelerated baseline, i.e., near-lossless visual quality with substantially reduced computation. Qualitative results in Figure~\ref{fig:video_vis} further show that TAP better preserves high consistency and fidelity, validating the effectiveness of token-adaptive predictor assignment for balancing efficiency and output quality.

\begin{figure}[t]
    \centering
    \begin{subfigure}[t]{0.48\linewidth}
        \centering
        \includegraphics[width=\linewidth]{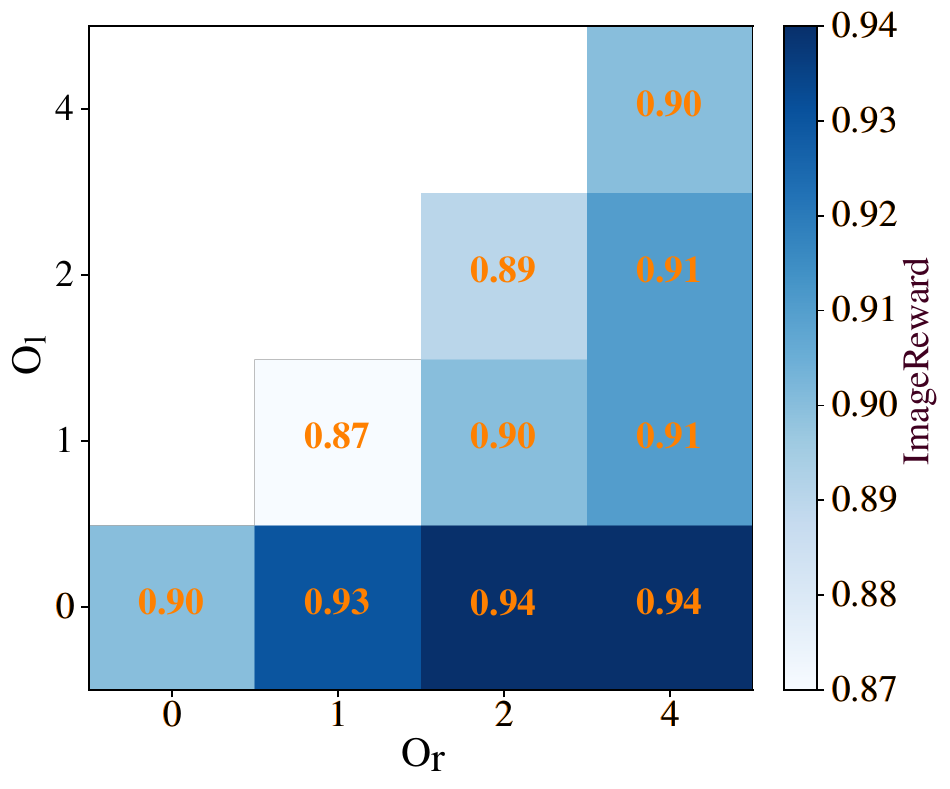}
        \caption{Effect of prediction order.}
        \label{fig:range_effect}
    \end{subfigure}
    \hfill
    \begin{subfigure}[t]{0.48\linewidth}
        \centering
        \includegraphics[width=\linewidth]{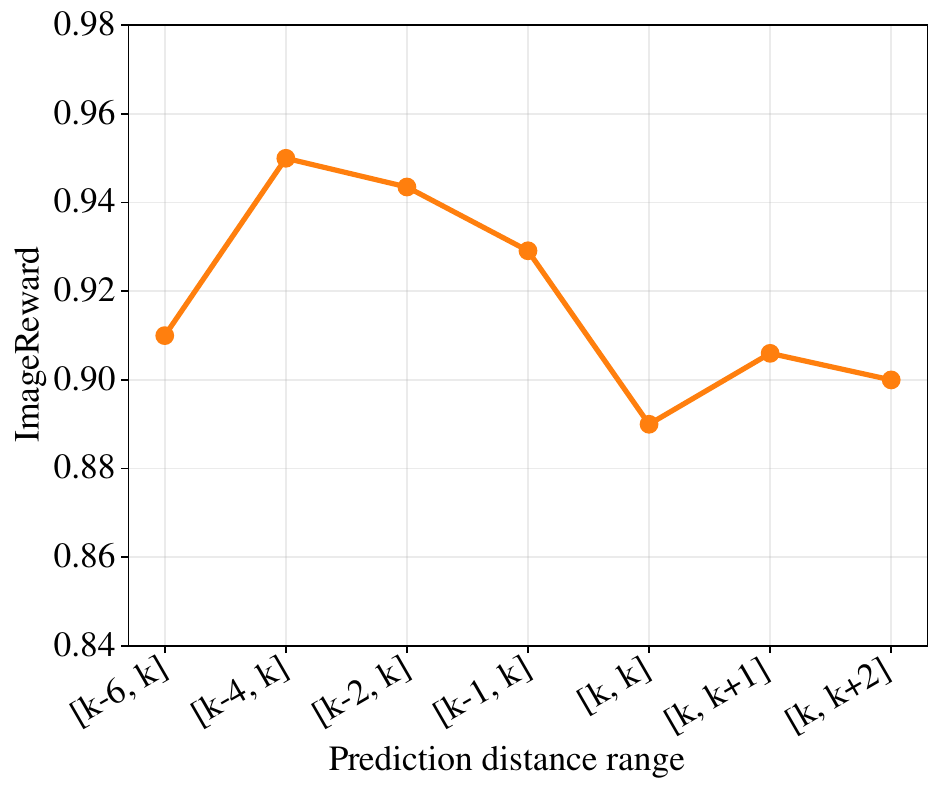}
        \caption{Effect of predictor distance.}
        \label{fig:order_effect}
    \end{subfigure}
    \vspace{-3pt}
    \caption{\textbf{Ablation on Taylor predictor family.} We investigate the influence of distance range and order configuration in TAP with $N=7$ on FLUX.1-dev. In (a), we fix the prediction distance as $k$, and in (b), we set the order as 2 and $\delta=1$.}
    \label{fig:ablation_taylor}
\end{figure}

\subsubsection{Ablation Study}
\paragraph{Ablation of the Taylor Predictor Family}
We construct a Taylor predictor family and study how predictor count and type affect generation quality. Concretely, we vary the prediction horizon (the discretized range $[k-\lambda, k]$) and the Taylor order range $[O_l,O_r]$, reporting results in Figure~\ref{fig:ablation_taylor}. Our ablation shows that increasing either the prediction distance or the expansion order yields substantial gains (raising a baseline ImageReward from about 0.89 ($O=2$) to 0.95). The best single setting in our sweep was $O_l=0$, $O_r=2$ with $\lambda=4$, and jointly varying both distance and order produced the largest improvement (from 0.95 to 0.99). Performance continues to improve as the predictor family grows but with diminishing returns and eventual saturation. Increasing granularity (e.g., using $\delta=0.1$ instead of $\delta=1$) produces only a small additional gain (about 0.005 ImageReward), so we use $\delta=1$ by default for simplicity.

Two additional observations emerge. First, including zeroth-order predictors (order $0$) is particularly valuable: they are more robust to abrupt, noncontinuous token dynamics and therefore complement higher-order predictors, yielding larger gains than using only high-order variants. Second, shifting the prediction window to the left (earlier expansion points) gives notable improvements because it avoids extrapolating beyond a token's Taylor convergence radius; moving the window to the right (e.g., $[k,k{+}2]$) yields little benefit. These findings validate the assumptions and effectiveness of our method.

\begin{figure*}[htbp]
\centering
\includegraphics[width=0.8\linewidth]{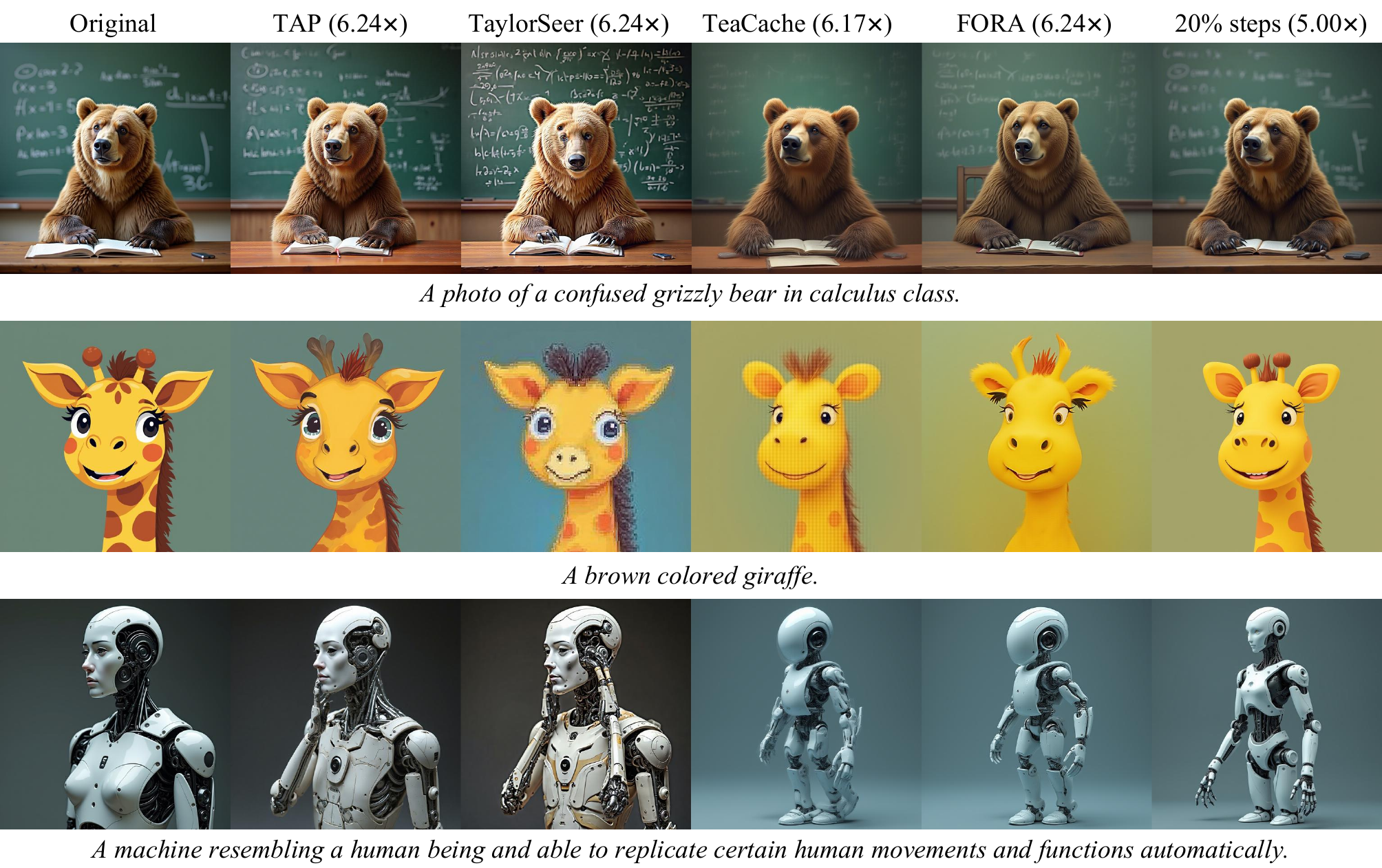}
\caption{Visualization results. On FLUX.1-dev, TAP delivers higher speedup without quality loss.}
\label{fig:visualization}
\vspace{-2mm}
\end{figure*}

\paragraph{Comparison with Global Predictors.}
We constructed a pool of 30 candidate predictors by systematically varying Taylor expansion order and prediction horizon, and evaluated each predictor's generative performance in isolation. The ImageReward of individual global predictors varied roughly between 0.86 and 0.92, while PSNR ranged approximately from 14.51 to 15.32, no single global predictor was best across all acceleration regimes. Instead, TAP's probe-driven, per-token selection adaptively fuses these diverse predictors and consistently outperforms any individual predictor, showing that the improvements arise from intelligently combining complementary predictors rather than from a single optimal sampler.

\subsection{Qualitative Analysis}
We present representative visualizations in Figure~\ref{fig:visualization} and Figure~\ref{fig:video_vis}. Across both image and video examples, cache-only and global-forecast baselines exhibit noticeable degradation at high acceleration ratios, manifesting as blurred details, distorted object geometry, and misalignment with text conditions. In contrast, TAP effectively preserves fine-grained textures, structural integrity, and visual consistency. This advantage comes from TAP’s token-adaptive assignment, which selects the best predictor for each token at every timestep and thus preserves perceptual fidelity even at high acceleration ratios. These qualitative observations match the quantitative gains reported above.

%% file: tables/1_flux.tex
\begin{table*}[hbt]
\centering
\caption{\textbf{Quantitative comparison in text-to-image generation} for FLUX.1-dev and FLUX.1-schnell (a distilled version). Best results are highlighted in \textbf{bold}, and second-best are \underline{underlined}.}
\vspace{-3mm}
\resizebox{\textwidth}{!}{
\begin{tabular}{l | c  c | c  c | c  c | c c c}
    \toprule
    \multirow{2}{*}{\textbf{Method}}
    & \multicolumn{4}{c|}{\textbf{Acceleration}} 
    & \multicolumn{2}{c|}{\textbf{Quality Metrics}} 
    & \multicolumn{3}{c}{\textbf{Perceptual Metrics}}\rule{0pt}{2ex}\\
    \cline{2-10}
    & \textbf{Latency(s) \(\downarrow\)} 
    & \textbf{Speed \(\uparrow\)} 
    & \textbf{FLOPs(T) \(\downarrow\)}  
    & \textbf{Speed \(\uparrow\)} 
    & \textbf{ImageReward\(\uparrow\)} 
    & \textbf{CLIP\(\uparrow\)} 
    & \textbf{PSNR\(\uparrow\)} 
    & \textbf{SSIM\(\uparrow\)}
    & \textbf{LPIPS\(\downarrow\)}\rule{0pt}{2ex}\\
    \midrule

    50 steps & 42.18 & 1.00$\times$ & 3726.87 & 1.00$\times$ & {0.95} & 30.63 & - &  - & -  \\
    40\% steps & 17.08 & 2.47$\times$ & 1490.75 & 2.50\(\times\) & \textbf{0.93} & \textbf{30.84} & \textbf{15.60} & \textbf{0.69} & \textbf{0.40} \\
    30\% steps & \underline{13.14} & \underline{3.21}$\times$ & \underline{1118.06} & \underline{3.33}\(\times\) & \underline{0.92} & \underline{30.70} & \underline{14.81} & \underline{0.66} & \underline{0.44} \\
    20\% steps & \textbf{8.52 }& \textbf{4.95}$\times$ & \textbf{745.37} & \textbf{5.00}\(\times\) & 0.89 & 30.28 & 13.96 & 0.62 & 0.50 \\
    \midrule
    FORA ($N=4$) & 12.37 & 3.41$\times$ & 969.93 & 3.84\(\times\) & 0.87 & 30.45 & 14.28 & 0.64 & 0.47 \\
    TeaCache ($l=1.1$) & 12.33 & 3.42$\times$ & 913.98 & 4.08\(\times\) & 0.93 & 30.47 & 15.19 & 0.65 & 0.47  \\
    TaylorSeer ($N=5, O=2$) & 12.02 & 3.51$\times$ & 895.32 & 4.16\(\times\) & {0.95} & 30.79 & \underline{16.73} & 0.71 & 0.36 \\
    SpeCa ($N=2, M=8$) & \textbf{11.68} & \textbf{3.61}$\times$ & \textbf{845.41} & \textbf{4.41}\(\times\) & \underline{0.96} & 30.73 & 16.09 & 0.68 & 0.40 \\
    \rowcolor{gray!20} 
    \textbf{TAP} ($N=5$) & \underline{11.82} & \underline{3.57}$\times$ & \underline{895.08} & \underline{4.16}\(\times\) & \textbf{0.97} & \textbf{30.86} & \textbf{17.43} & \textbf{0.71} & \textbf{0.34} \\
    \midrule
    FORA ($N=5$) & \underline{10.12} & \underline{4.17}$\times$ & 746.29 & 4.99\(\times\) & 0.89 & 30.55 & 13.82 & 0.61 & 0.50 \\
     TeaCache ($l=1.4$)  & 11.16 & 3.78$\times$ & 779.83 & 4.77\(\times\) & 0.92 & 30.44 &  14.79 & 0.63 & 0.51 \\
    
    TaylorSeer ($N=6, O=2$) & \textbf{10.02} & \textbf{4.21}$\times$ & \underline{746.29} & \underline{4.99}\(\times\) & \underline{0.94} & \underline{30.76} & \underline{15.88} & \underline{0.67} & \underline{0.42} \\
    SpeCa ($N=4, M=8$) & 10.76 & 3.92$\times$ & 755.32 & 4.93\(\times\) & 0.93 & 30.57 & 15.72 & 0.66 & 0.43 \\
    \rowcolor{gray!20}
    \textbf{TAP} ($N=6$) & {10.21} & 4.13$\times$ & \textbf{746.02} & \textbf{5.55\(\times\)} & \textbf{0.96} & \textbf{30.92} & \textbf{16.76} & \textbf{0.68} & \textbf{0.39} \\
    \midrule
    FORA ($N=7$) & \underline{7.91} & \underline{5.33}$\times$ & 597.26 & {6.24\(\times\)} & 0.80 & 30.42 & 13.43 & 0.60 & 0.55 \\
    TeaCache ($l=2.0$)  & 8.59 & 4.91$\times$ & 604.02 & 6.17\(\times\) & 0.66 & 30.07 & 14.23 & 0.60 & 0.58 \\
    TaylorSeer ($N=8, O=2$) & 8.17 & {5.16}$\times$ & 597.26 & {6.24\(\times\)} & 0.91 & 30.62 & 14.72 & 0.61 & 0.50 \\
    \rowcolor{gray!20} 
    \textbf{TAP} ($N=8$) & 8.40 & 5.02$\times$ & \underline{596.99} & \underline{6.24\(\times\)} & \textbf{0.99} & \textbf{31.19} & \textbf{16.11} & \textbf{0.64} & \textbf{0.44} \\
    \rowcolor{gray!20} 
    \textbf{TAP} ($N=10$) & \textbf{6.21} & \textbf{6.79}$\times$ & \textbf{522.48} & \textbf{7.13\(\times\)} & \underline{0.91} & \underline{30.69} & \underline{15.94} & \underline{0.63} & \underline{0.49} \\

    \midrule
    FLUX-Schnell (4 steps) & 4.46 & 1.00$\times$ & 278.41 & 1.00$\times$ & 0.88 & 32.03 &  - &  - & -   \\
    TaylorSeer ($N=3, O=2$) & \underline{2.61} & \underline{1.71}$\times$ & \underline{139.23} & \underline{2.00\(\times\)} & \underline{0.86} & \underline{32.08} & \underline{22.67} & 0.72 & \underline{0.32} \\
    \rowcolor{gray!20} 
    \textbf{TAP} ($N=3$) & \textbf{2.55} & \textbf{1.75}$\times$ & \textbf{139.12} & \textbf{2.00\(\times\)} & \textbf{0.88} & \textbf{32.36} & \textbf{26.26} & \textbf{0.86} & \textbf{0.15} \\

    \bottomrule
\end{tabular}}
\label{table:FLUX-Metrics}
\raggedright
\end{table*}

%% file: tables/4_qwen_image.tex
\begin{table*}[htb]
\centering
\caption{\textbf{Quantitative comparison in text-to-image generation} for Qwen-Image and Qwen-Image-Lightning (a distilled version). Best results are highlighted in \textbf{bold}, and second-best are \underline{underlined}.}
\vspace{-3mm}
\resizebox{\textwidth}{!}{
\begin{tabular}{l | c c c | c c | c c c}
\toprule
\multirow{2}{*}{\textbf{Method}}
 & \multicolumn{3}{c|}{\textbf{Acceleration}}
 & \multicolumn{2}{c|}{\textbf{Quality Metrics}}
 & \multicolumn{3}{c}{\textbf{Perceptual Metrics}} \\
\cline{2-9}
 & \textbf{Latency (s) $\downarrow$}
 & \textbf{FLOPs (T) $\downarrow$}
 & \textbf{Speed $\uparrow$}
 & \textbf{ImageReward $\uparrow$}
 & \textbf{CLIP $\uparrow$}
 & \textbf{PSNR $\uparrow$}
 & \textbf{SSIM $\uparrow$}
 & \textbf{LPIPS $\downarrow$} \\
\midrule

50 steps
 & 127.10 & 12917.56 & - & 1.23 & 33.74 & - & - & - \\

50\% steps
 & \underline{63.54} & \underline{6458.78} & \underline{2.00}$\times$ & \textbf{1.20} & \textbf{33.77} & \textbf{18.35} & \textbf{0.77} & \textbf{0.28} \\

20\% steps
 & \textbf{25.54} & \textbf{2583.51} & \textbf{5.00}$\times$ & \underline{0.73} & \underline{32.24} & \underline{13.45} & \underline{0.57} & \underline{0.57} \\

\midrule

FORA ($N=3$)
 & 48.59 & 4392.93 & 2.94$\times$ & 0.92 & 32.25 & 14.12 & 0.59 & 0.50 \\

TeaCache ($l=0.8$)
 & 52.64 & 3621.18 & {3.57}$\times$ & \underline{1.18} & \underline{33.52} & \underline{19.07} & \underline{0.79} & \underline{0.27} \\

TaylorSeer ($N=4, O=2$)
 & \underline{44.37} & \underline{3617.96} & \underline{3.57}$\times$ & 1.18 & 33.44 & 18.02 & 0.76 & 0.30 \\
\rowcolor{gray!20}
TAP ($N=4$)
 & \textbf{40.96} & \textbf{3617.43} & \textbf{3.57}$\times$ & \textbf{1.23} & \textbf{33.80} & \textbf{20.13} & \textbf{0.81} & \textbf{0.22} \\

\midrule

FORA ($N=4$)
 & \underline{38.41} & 3359.99 & 3.84$\times$ & 0.70 & 31.98 & 13.13 & 0.53 & 0.58 \\

TeaCache ($l=1.0$)
 & 47.84 & 3276.45 & 3.94$\times$ & \underline{1.16} & \underline{33.47} & \underline{18.13} & \underline{0.73} & \underline{0.34} \\

TaylorSeer ($N=5, O=2$)
 & 40.70 & \underline{3101.30} & \underline{4.17$\times$} & 1.16 & 33.37 & 16.18 & 0.70 & 0.39 \\
\rowcolor{gray!20}
TAP ($N=5$)
 & \textbf{34.21} & \textbf{3100.76} & \textbf{4.17$\times$} & \textbf{1.19} & \textbf{33.55} & \textbf{18.72} & \textbf{0.78} & \textbf{0.27} \\

\midrule

Qwen-Image-Lightning (8 steps)
 & 7.46 & 560.96 & - & {1.30} & 33.06 & - & - & - \\

TaylorSeer ($N=5, O=2$) 
 & \textbf{4.68} & \underline{280.54} & \underline{2.00}$\times$ & \underline{1.21} &  \underline{33.03} &  \underline{17.59} &  \underline{0.70} &  \underline{0.27} \\
\rowcolor{gray!20}
TAP ($N=5$)
 & \underline{4.80} & \textbf{280.50} &  \textbf{2.00}$\times$ & \textbf{1.27} & \textbf{33.50} & \textbf{21.11} & \textbf{0.77} & \textbf{0.23} \\

\bottomrule
\end{tabular}
}
\label{table:qwen-image-Metrics}
\end{table*}

%% file: tables/7_hunyuan.tex
\begin{table}[htbp]
\centering
\caption{Quantitative comparison on VBench (HunyuanVideo).}
\setlength\tabcolsep{8pt}
\small
\resizebox{\linewidth}{!}{
\begin{tabular}{l c c c}
    \toprule
    {\bf Method} & {\bf Latency (s) $\downarrow$} & {\bf FLOPs (T)} & {\bf VBench (\%) $\uparrow$} \\
    \midrule
    50 steps & 121.32 & 5481.50\(_{(1.00\times)}\) & 66.61 \\
    50\% steps & \textbf{60.96}  & \textbf{2740.75}\(_{(2.00\times)}\) & \underline{66.38} \\
    20\% steps & \underline{24.80}  & \underline{1096.30}\(_{(5.00\times)}\) & \textbf{64.16} \\
    \midrule
    FORA ($N=5$) & \underline{31.68} & 1101.72\(_{(4.98\times)}\) & 63.87 \\
    TeaCache ($\ell=0.4$) & 32.33 & 1205.42\(_{(4.55\times)}\) & \underline{65.13} \\
    TeaCache ($\ell=0.5$) & \textbf{25.53} & \textbf{991.96\(_{(5.53\times)}\) } & 64.28 \\
    Taylor ($N=6, O=2$) & 33.67 & 1101.72\(_{(4.98\times)}\) & 64.89 \\
    TAP ($N=6$) & {31.91} & \underline{1101.47}\(_{(4.98\times)}\) & \textbf{65.46} \\
    \bottomrule
\end{tabular}
}
\label{table:HunyuanVideo-Metrics}
\end{table}

%% file: sec/5_conclusion.tex
\section{Conclusion}
We present TAP, a probe-driven, token-adaptive diffusion-acceleration framework that makes per-token predictions based on a lightweight proxy. TAP is highly efficient, fully parallelizable, and compatible with a wide range of predictor designs. It achieves substantial speedups in diffusion sampling with minimal memory and compute overhead while preserving perceptual quality. Experiments demonstrate consistent gains across both image and video models.

%% file: sec/X_suppl.tex
\clearpage
\setcounter{page}{1}
\setcounter{figure}{4}
\setcounter{table}{3}
\maketitlesupplementary

\section{Implementation Details}

\paragraph{Model Specification}
The images generated by FLUX.1-dev, FLUX.1-schnell, and Qwen-Image-Lightning were produced at a resolution of $1024\times1024$, while outputs from Qwen-Image were produced at $1328\times1328$. Video frames from HunyuanVideo were obtained at $480\times832$.  
All reported measurements (performance, FLOPs, latency, and memory usage) are normalized to per-sample values at a batch size of 1.
All metrics were collected on identical hardware across the evaluation dataset (e.g., 200 trials on DrawBench), with 3 additional warm-up inference runs performed before measurement.

\paragraph{FLOPs and HBM}
We measure FLOPs using the open-source tool \texttt{calflops}.
For a Transformer with $L$ layers, sequence length $N_x$ and hidden dimension $D$, a compact estimate is $\mathrm{FLOPs}\approx L\big(24N_xD^{2}+4N_x^{2}D\big)$, where the $24N_xD^{2}$ term approximates linear projections and FFN matmuls and the $4N_x^{2}D$ term captures attention matmuls. During reuse steps, only the modulated input layer (Eq.~(5)) is recomputed, which reduces the overall FLOPs.
Peak HBM grows roughly linearly with batch $B$ and tokens $N_x$, a simple upper bound is $\mathrm{HBM}_{\mathrm{peak}}(B,N_x)\approx P\cdot b + (|C|+\alpha)\,B\,N_x\,D\cdot b$, where $P$ is the parameter count, $b$ is bytes per element (e.g. $b=2$ for fp16), $|C|$ is the number of cached feature tensors (shallow probe and residual features), and $\alpha$ models transient activations. Prior caching methods require storing per-layer features, whereas TAP only caches the probe and residual features, thereby substantially reducing memory requirements for caching.

\paragraph{Baseline Specification}
We implement FORA, TeaCache, TaylorSeer, and SpeCa using their official codebases. In FORA, TaylorSeer, and TAP, $N$ denotes the cache-window length: we compute on the first step and skip the remaining $N-1$ steps within each window of $N$ steps. In TaylorSeer, $O$ denotes the maximum prediction order. In TeaCache, $l$ is the threshold used to decide whether to compute or reuse cached features at the beginning of each step. In SpeCa, $N$ and $M$ denote the minimum and maximum caching steps, respectively, for each sample. Following prior works \cite{liuTaylorSeer2025, liu2025speca}, we retain the features computed in the previous three timesteps during sampling (two timesteps for FLUX.1-schnell), since preserving early computations is important for generation quality. The same protocol is applied to the TaylorSeer and SpeCa baselines. Besides, we use the official flow-matching sampler for step-reduced baselines (same setup across comparisons).

\section{Differences from Existing Work}
Our approach differs from prior training-free acceleration methods in several important ways:

\paragraph{Comparison to Direct-Reuse Methods}
Unlike direct feature-reuse methods~\cite{selvaraju2024fora, yuan2024ditfastattn, ma2024deepcache}, TAP leverages predictor-based forecasting to aggregate information from multiple past fully-computed steps and predict future features. Note that direct reuse is a special case of our predictor family (equivalent to a zero-order predictor), so TAP can exploit both direct-reuse and prediction-based strategies jointly.

\paragraph{Comparison to Prediction-based Methods}
Existing predictor-based accelerators such as TaylorSeer~\cite{liuTaylorSeer2025} and ABCache~\cite{yu2025ab} typically rely on a single, fixed predictor. By contrast, TAP dynamically and organically combines multiple predictor types per token and per timestep, which improves generation quality. Methods that split features into frequency bands and apply fixed predictors per band (e.g., FreqCa~\cite{liu2025freqca}) are still limited by their use of fixed predictors. TAP’s more diverse candidate set captures a wider range of temporal behaviors.

\paragraph{Comparison to Sample-Adaptive Methods}
Sample-adaptive methods like SpeCa~\cite{liu2025speca} and TeaCache~\cite{liu2024timestep} allocate computation on a per-sample basis by deciding whether to compute or skip each step. While sample-adaptive strategies are powerful, they often break efficient batch-parallel inference due to varying per-sample workloads and typically require hand-tuned thresholds that must be re-tuned across models (e.g., TeaCache’s polynomial coefficients). TAP is threshold-free and supports batch-parallel inference without manual threshold tuning. Besides, it can incorporate sample-adaptive ideas. For example, falling back to full computation when proxy losses are large.

\paragraph{Comparison to TaylorSeer}
TaylorSeer is an influential prediction-based method. We extend its insights in two ways. First, we observe that prediction behavior varies substantially across predictor order and distance, so we construct a Taylor predictor family to better capture per-token temporal heterogeneity. Second, TAP can benefit from predictors beyond Taylor expansions (see Table~\ref{tab:alt-predictors}), demonstrating broader compatibility and improved robustness.

\paragraph{Summary}
In short, TAP unifies the strengths of existing approaches while avoiding their main limitations (fixed predictors, hand-tuned thresholds, or loss of batch parallelism). Empirically, TAP results in substantial improvements in generation quality and efficiency compared to prior methods.

\section{More Quantitative Experiments}

\begin{figure}[t]
    \centering
    \begin{subfigure}[t]{0.48\linewidth}
        \centering
        \includegraphics[width=\linewidth]{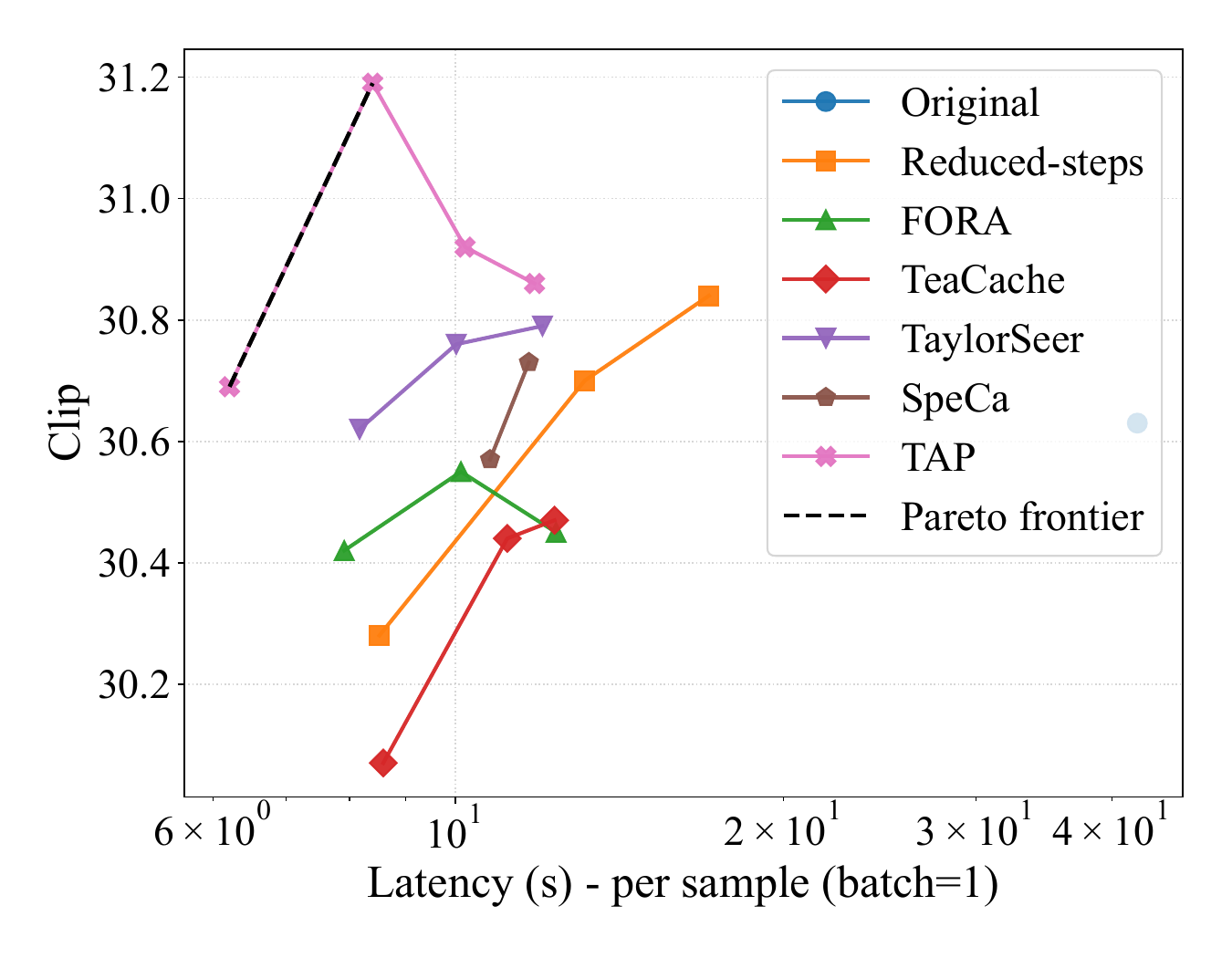}
    \end{subfigure}
    \hfill
    \begin{subfigure}[t]{0.48\linewidth}
        \centering
        \includegraphics[width=\linewidth]{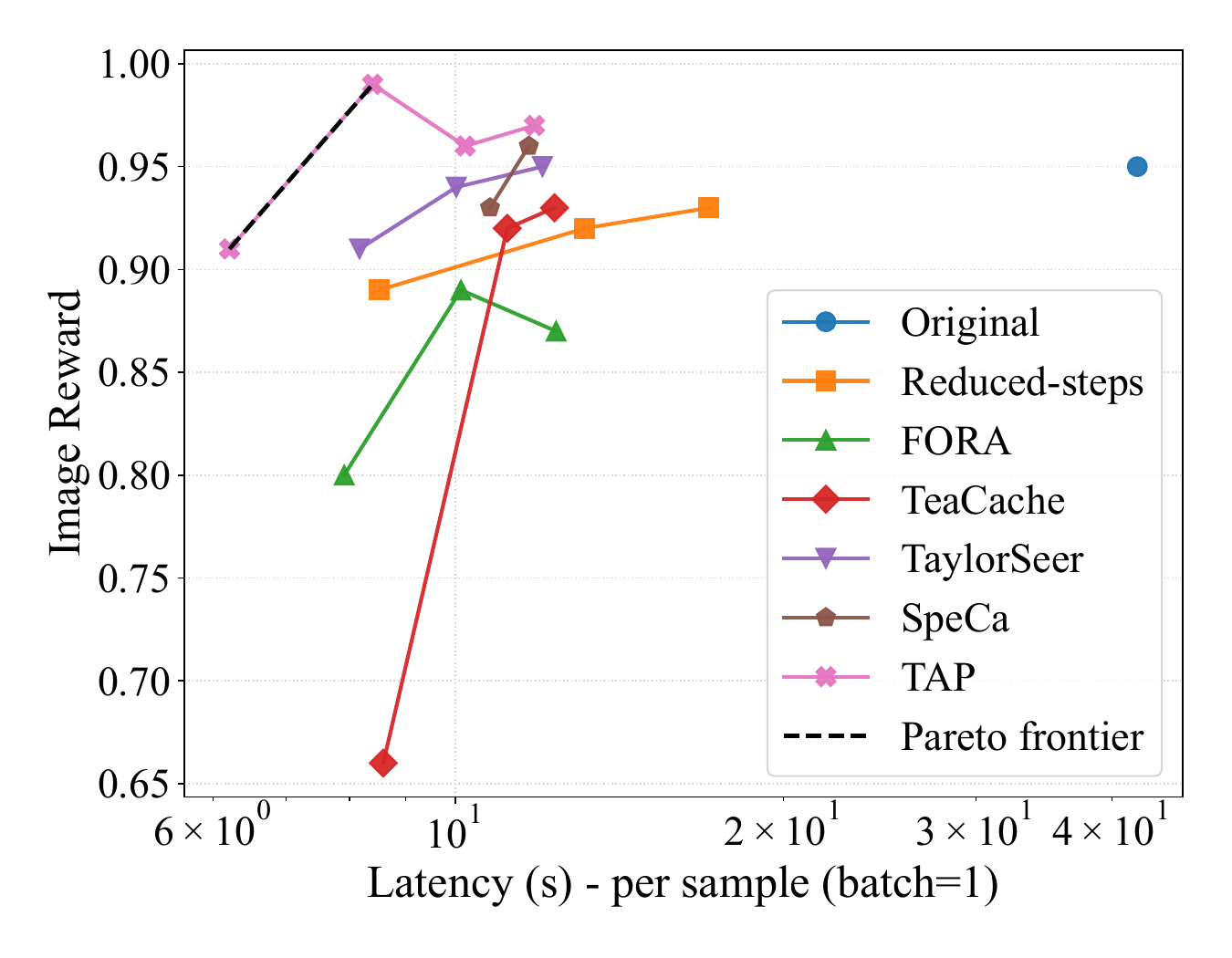}
    \end{subfigure}
    \vspace{-8pt}
    \caption{Pareto plots on FLUX.1-dev evaluated on DrawBench.}
    \label{fig:pareto_figures}
    \vspace{-10pt}
\end{figure}

\paragraph{Pareto Plots} 
To better illustrate the speed–quality trade-off, we provide Pareto plots. As shown in Figure \ref{fig:pareto_figures}, we present latency vs. CLIP and latency vs. ImageReward curves on FLUX.1-dev to characterize the speed–quality frontier.

\paragraph{Summary of Hyperparameter Sensitivity}
We provide a comprehensive set of ablation experiments to justify our default hyperparameter choices. In particular, Figure~2 presents systematic sweeps over predictor order \(M\) and prediction distance \(\lambda\), demonstrating the typical diminishing returns as order or prediction distance grows and identifying a stable operating region that balances accuracy and cost. We also evaluate the effect of selection granularity \(\delta\), as noted in the manuscript, ``Increasing granularity (e.g., using \(\delta = 0.1\) instead of \(\delta = 1\)) produces only a small additional gain (about 0.005 ImageReward), so we use \(\delta = 1\) by default for simplicity.'' Finally, the cache window / acceleration control parameter \(N\) is swept in Tables~1--3, which report results for multiple values of \(N\) and thereby quantify the expected fidelity--speed tradeoff. Taken together, these experiments show that (i) moderate predictor orders and short-to-medium prediction distances yield the best trade-off in practice, (ii) a coarse granularity \(\delta=1\) is a robust and simple default because finer granularity brings only marginal improvements, and (iii) \(N\) cleanly controls the acceleration ratio with predictable effects on quality as documented in Tables~1--3.

\paragraph{Alternative Predictors}
We introduce the Taylor predictor family to capture diverse temporal dynamics, and emphasize that TAP is compatible with other predictor families \cite{liu2025freqca, zhengFoCa2025}. To demonstrate this compatibility, we additionally evaluate a nonlinear predictor built from probabilists' Hermite polynomials~\cite{fengHiCache2025, liu2025freqca} with second-order prediction ($m=2$). For a normalized time $t\in[-1,1]$, the predicted feature \(\widehat{\mathbf{r}}_{t}\) for token \(i\) is modeled as
\[
    \widehat{\mathbf{r}}_{t}^{(i)} \;=\; \sum_{k=0}^{m} c_{i,k}\,\mathrm{He}_{k}(t),
\]
where \(\{\mathrm{He}_{k}\}_{k=0}^{m}\) are Hermite polynomials and the coefficients \(c_{i,k}\) are estimated by ordinary least-squares using the \(K\) most recent cached steps. We form a design matrix from the recent timestamps and solve for \(c_{i,\cdot}\) that minimize the squared prediction error on those cached features.

The results in Table~\ref{tab:alt-predictors} show that combining TAP with the Hermite-based predictor yields additional gains beyond TAP alone, confirming that our framework is compatible with different feature-prediction methods and can benefit from richer predictor families.

\begin{table}[h]
\centering
\caption{Comparison of alternative predictors. Experiments conducted on the FLUX.1-dev baseline; TAP with \(N=5\).}
\label{tab:alt-predictors}
\small
\begin{tabular}{lcc}
\toprule
Method & ImageReward\(\uparrow\) & PSNR\(\uparrow\) \\
\midrule
Baseline & 0.95 & 16.69  \\
TAP & \underline{0.97} & \underline{17.43} \\
TAP (w/ Hermite) & \textbf{0.98} & \textbf{17.52} \\
\bottomrule
\end{tabular}
\end{table}

\paragraph{Memory Usage Analysis}
We report peak GPU memory usage for each backbone in Table~\ref{tab:memory}. TAP introduces only \(0.29\,\mathrm{GB}\) of additional memory (approximately \(0.6\%\) of the original model), whereas TaylorSeer requires \(11.91\,\mathrm{GB}\) of extra memory (about \(26.27\%\)) on HunyuanVideo model. This demonstrates that TAP imposes negligible memory overhead while still providing substantial acceleration.

\begin{table}[h]
\centering
\caption{Peak GPU memory usage (GB) during sampling. }
\label{tab:memory}
\small
\begin{tabular}{lccc}
\toprule
Backbone & Original & TaylorSeer  & TAP   \\
\midrule
FLUX.1-dev        & 37.72 & 45.17 & 37.82  \\
FLUX.1-schnell    & 37.66 & 44.80 & 37.80  \\
Qwen-Image        & 65.21 & 96.03 & 65.72  \\
Qwen-Image-Lightning  & 63.04 & 69.26 & 63.14  \\
HunyuanVideo     & 45.33 & 57.24 & 45.62  \\
\bottomrule
\end{tabular}
\end{table}

\paragraph{Choice of Probe}
Our method uses the timestep-embedding–modulated input as the probe for computing proxy losses. For comparison, we also evaluate the raw model input. Results are reported in Table~\ref{tab:probe_choice}. We find that the modulated input yields larger gains than the raw input. We attribute this to the stronger correlation between the modulated input and the model output dynamics (i.e., the modulated input better reflects how features evolve over timesteps), which is consistent with the analysis in Figure~\ref{fig:probe-corr}.

\begin{table}[h]
\centering
\caption{Comparison of probes. Experiments conducted on the FLUX.1-dev baseline; TAP with \(N=8\).}
\label{tab:probe_choice}
\small
\begin{tabular}{lcc}
\toprule
Method & ImageReward\(\uparrow\) & PSNR\(\uparrow\) \\
\midrule
Baseline & 0.90 & 14.70 \\
TAP (input probe) & \underline{0.97} & \underline{15.98} \\
TAP (modulated-input probe) & \textbf{0.99} & \textbf{16.11} \\
\bottomrule
\end{tabular}
\end{table}

\paragraph{Comparison of Proxy Losses}
We compare the cosine distance against the $\ell_1$ loss as a proxy loss to quantify prediction errors. 
Results are presented in Table~\ref{tab:proxy_losses}. Both losses deliver clear improvements over existing baselines, with the cosine proxy showing a stronger advantage compared to the $\ell_1$ in our experiments.

\begin{table}[htbp]
\centering
\caption{Comparison of proxy losses. Experiments conducted on the FLUX.1-dev baseline; TAP with \(N=8\).}
\label{tab:proxy_losses}
\small
\begin{tabular}{lcc}
\toprule
Method & ImageReward\(\uparrow\) & PSNR\(\uparrow\) \\
\midrule
Baseline & 0.90 & 14.70 \\
TAP ($\ell_1$ loss) & \underline{0.97} & \underline{16.05} \\
TAP (cosine loss) & \textbf{0.99} & \textbf{16.11} \\
\bottomrule
\end{tabular}
\end{table}

\begin{figure}[htbp]
\centering
\includegraphics[width=0.9\linewidth]{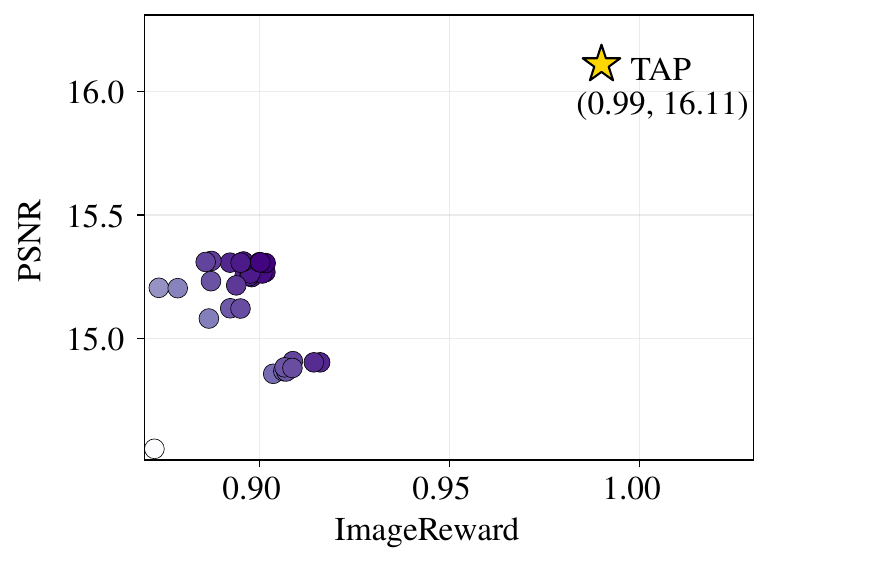}
\caption{Comparison to global predictors. Experiments are conducted on FLUX.1-dev with 50 sampling steps and a fixed acceleration of \(N=8\). TAP achieves better performance than a single global predictor.}
\label{fig:global_predictor}
\vspace{-2mm}
\end{figure}

\paragraph{More Comparison with Global Predictors}
In the main manuscript we compare a single global predictor with our token-adaptive predictor. Then we further illustrate those results in Figure~\ref{fig:global_predictor}. As shown, TAP clearly outperforms a single global predictor under the same sampling budget and acceleration settings, demonstrating the benefit of per-token adaptation.

\paragraph{Human Evaluation} 
We further include FID, a VQA score, and a human preference study for comprehensive evaluation, with all results reported in Table~\ref{tab:fid_vqa_userstudy_example}.

\begin{table}[ht]
\centering
\caption{With FLUX.1-dev on DrawBench, FID is computed between images from accelerated and original (unaccelerated) models. Preference is the fraction of sample-level judgments favoring the method, averaged over 10 human raters.}
\footnotesize
\begin{tabular}{l c c c }
\toprule
Method & FID $\downarrow$ & VQAScore (\%) $\uparrow$ & Preference (\%) $\uparrow$ \\
\midrule
TaylorSeer ($N$=8) & 102.25 & 77.43 & 16.5  \\
TAP ($N$=8)        & \textbf{82.98}  & \textbf{77.90} & \textbf{35.5} \\
\bottomrule
\end{tabular}
\label{tab:fid_vqa_userstudy_example}
\end{table}

\section{More Qualitative Experiments}
\begin{figure}[htbp]
\centering
\includegraphics[width=0.95\linewidth]{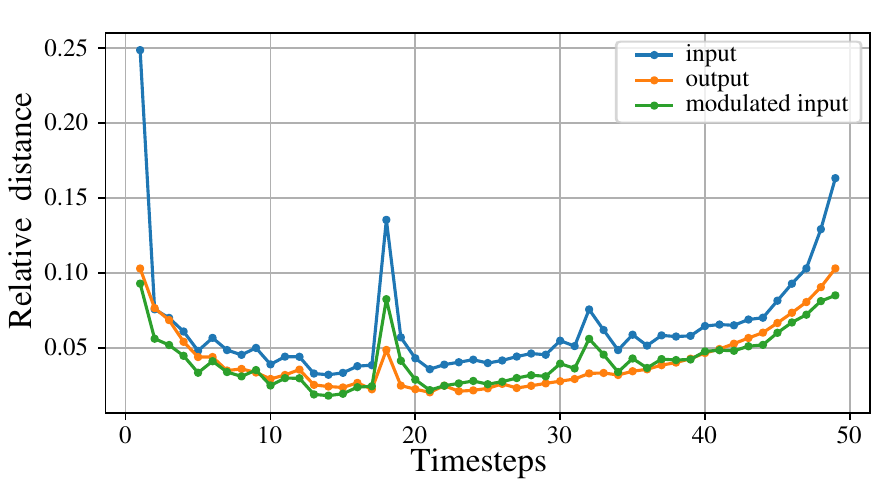}
\caption{Correlation between probe proxy loss and downstream feature error. Strong positive correlation supports the probe-then-select design.}
\label{fig:probe-corr}
\vspace{-2mm}
\end{figure}

\paragraph{Input-Output Analysis}
Our method estimates predictor performance from the prediction error of the model input, based on the strong relationship between a model's input and its output as shown in prior work \cite{liu2024timestep, bu2025dicache}. We visualize the differences of the input, the modulated input, and the model output between consecutive timesteps in Figure~\ref{fig:probe-corr}. The results show that both the raw input and the modulated input exhibit a strong correlation with the model output. In particular, the modulated input shows a stronger correlation, thus we choose it as our probe.

\paragraph{Token Heterogeneities Analysis}
We additionally analyze the relative error of token features over time. As shown in Figure~\ref{fig:token_heterogeneous}, different tokens can exhibit markedly different temporal dynamics: for example, low-texture regions tend to vary smoothly across timesteps, while high-texture or content-rich regions may undergo abrupt changes at mid timesteps during generation. These observations motivate adaptive, per-token assignment of predictors to better match each token's temporal behavior.

\begin{figure}[h]
  \centering
  \includegraphics[width=1.0\linewidth]{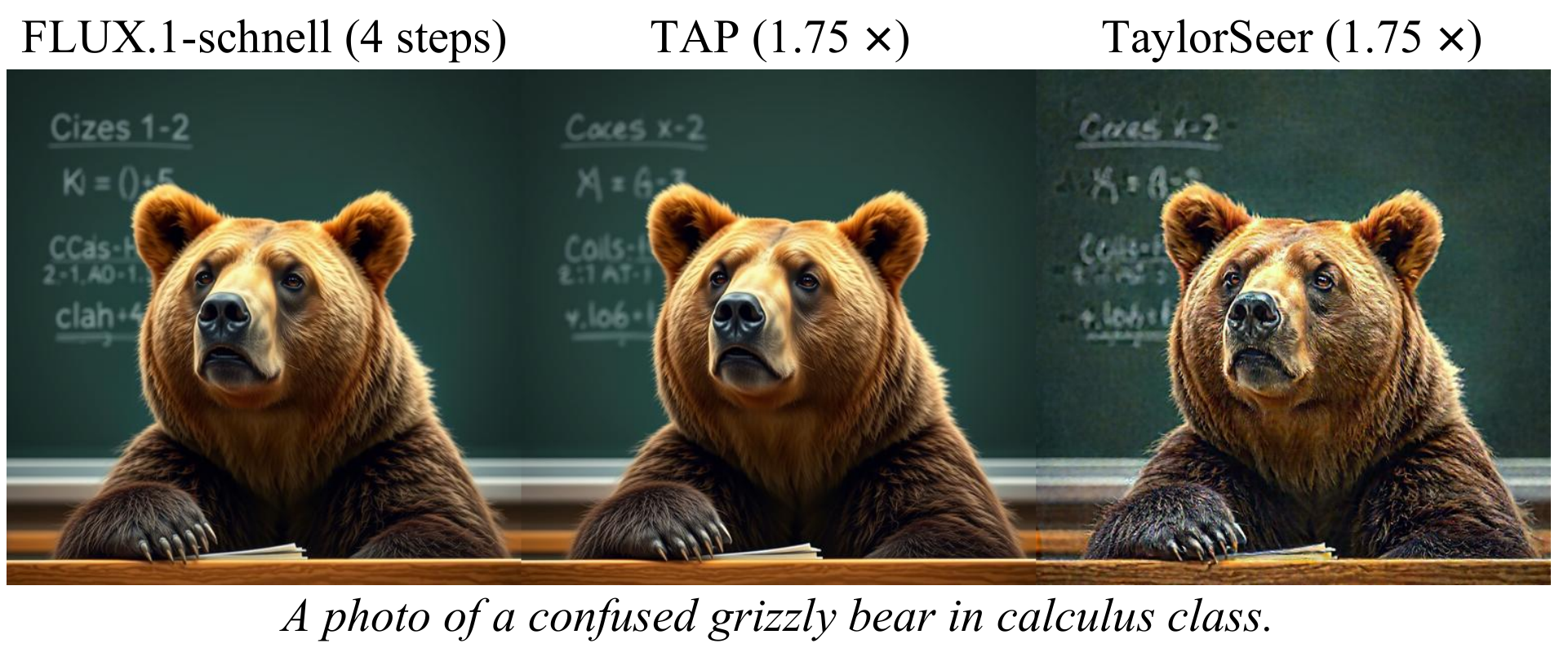}
  \caption{Visualization results on FLUX.1-schnell with 4 steps.}
  \label{fig:vis_flux-schnell}
\end{figure}

\begin{figure*}[htbp]
    \centering
    \begin{subfigure}[t]{0.29\linewidth}
        \centering
        \includegraphics[width=\linewidth]{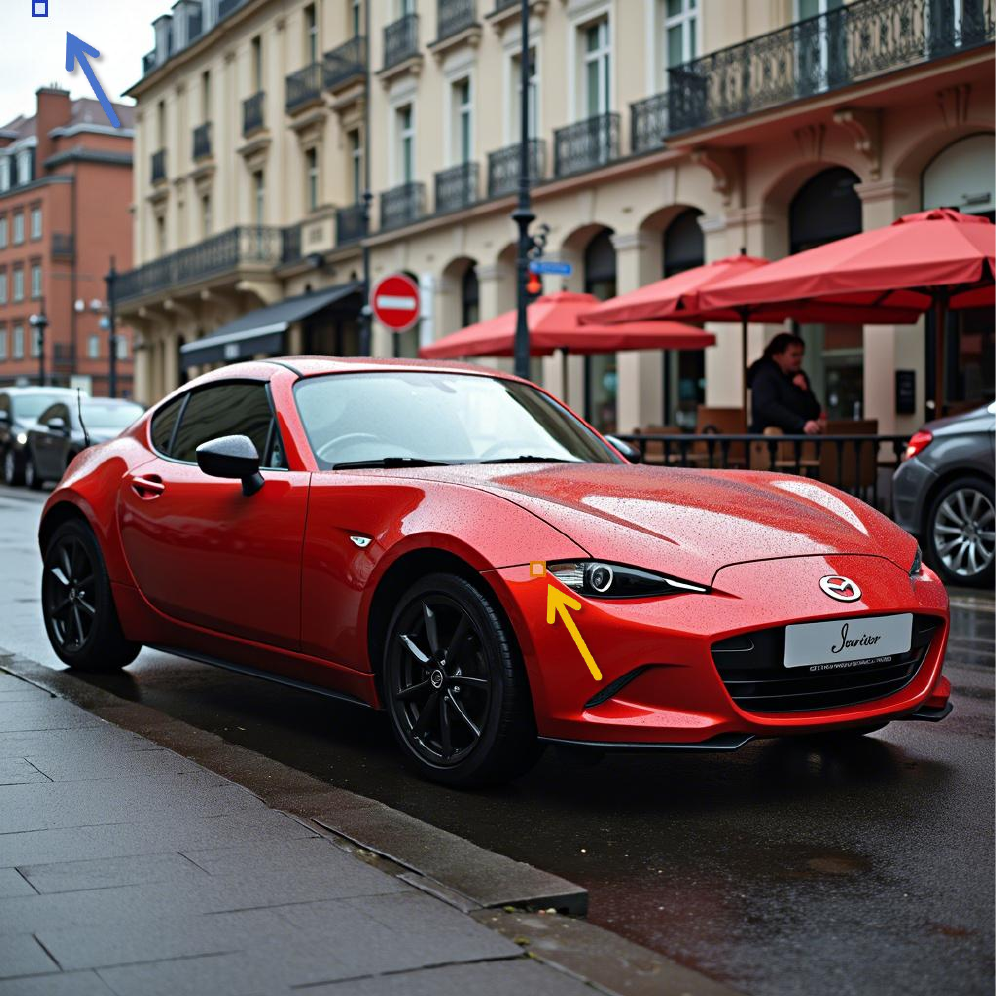}
        \caption{Generation example.}
        \label{fig:token_example}
    \end{subfigure}
    \hfill
    \begin{subfigure}[t]{0.48\linewidth}
        \centering
        \includegraphics[width=\linewidth]{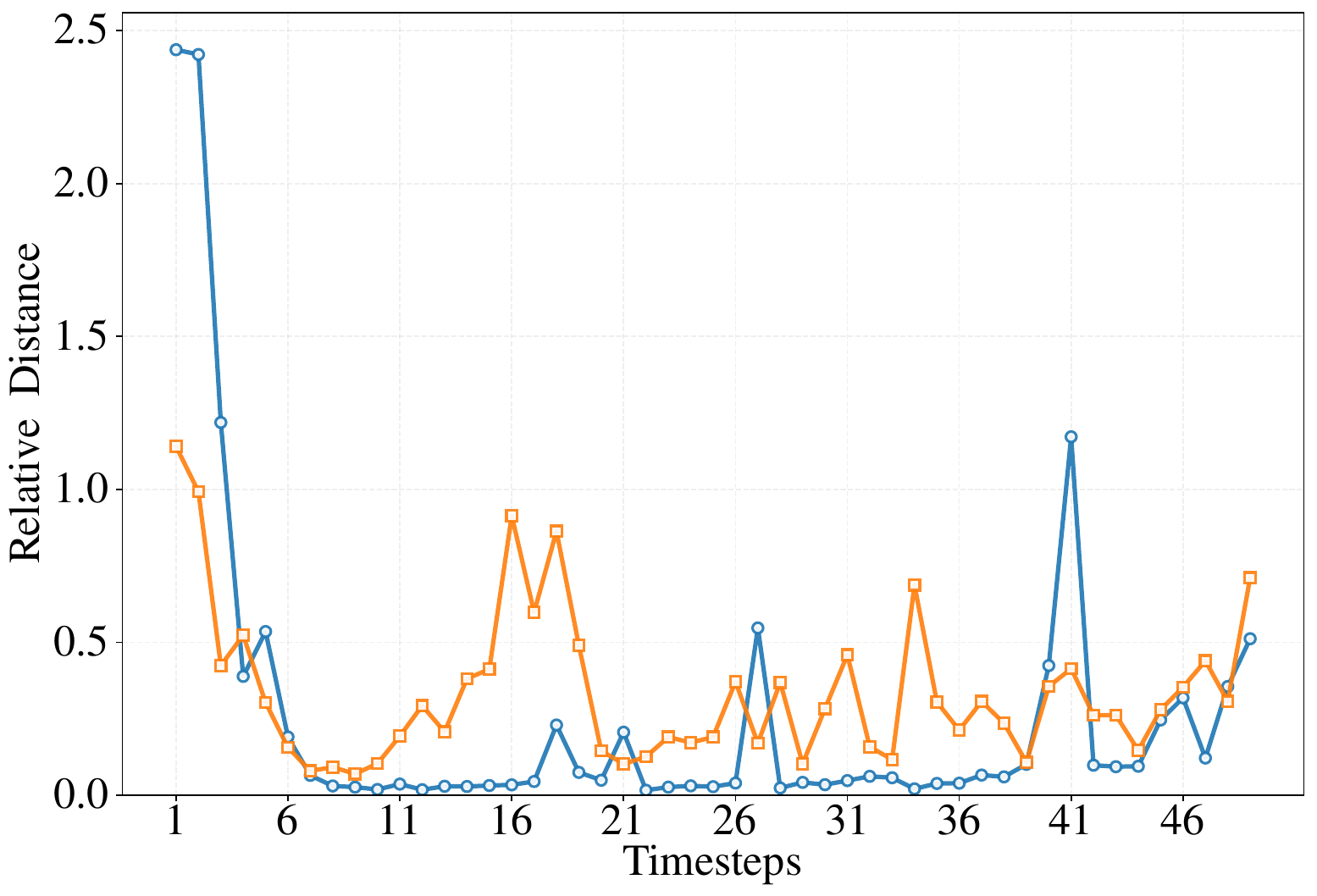}
        \caption{Token distance across adjacent timesteps.}
        \label{fig:token_similarity}
    \end{subfigure}
    \vspace{-3pt}
    \caption{Analysis of token evolution patterns. Different tokens follow different temporal patterns: low-texture regions (blue token) change smoothly across timesteps, whereas content-rich regions (orange token) exhibit discontinuous and larger variations around mid timesteps. Results were obtained on FLUX.1-dev with 50 sampling steps.}
    \label{fig:token_heterogeneous}
\end{figure*}

\begin{figure*}[h]
    \centering
    \begin{subfigure}[h]{0.48\linewidth}
        \centering
        \includegraphics[width=\linewidth]{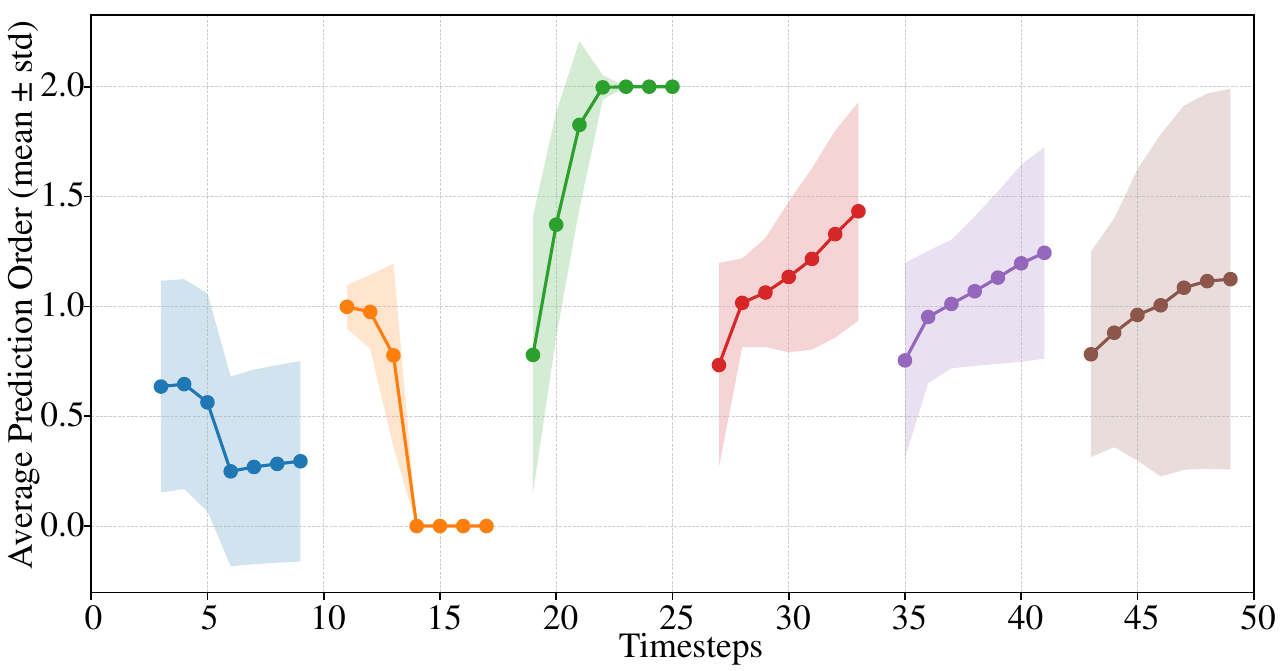}
        \caption{Average selected prediction order over timesteps.}
        \label{fig:avg_order_plot}
    \end{subfigure}
    \hfill
    \begin{subfigure}[h]{0.48\linewidth}
        \centering
        \includegraphics[width=\linewidth]{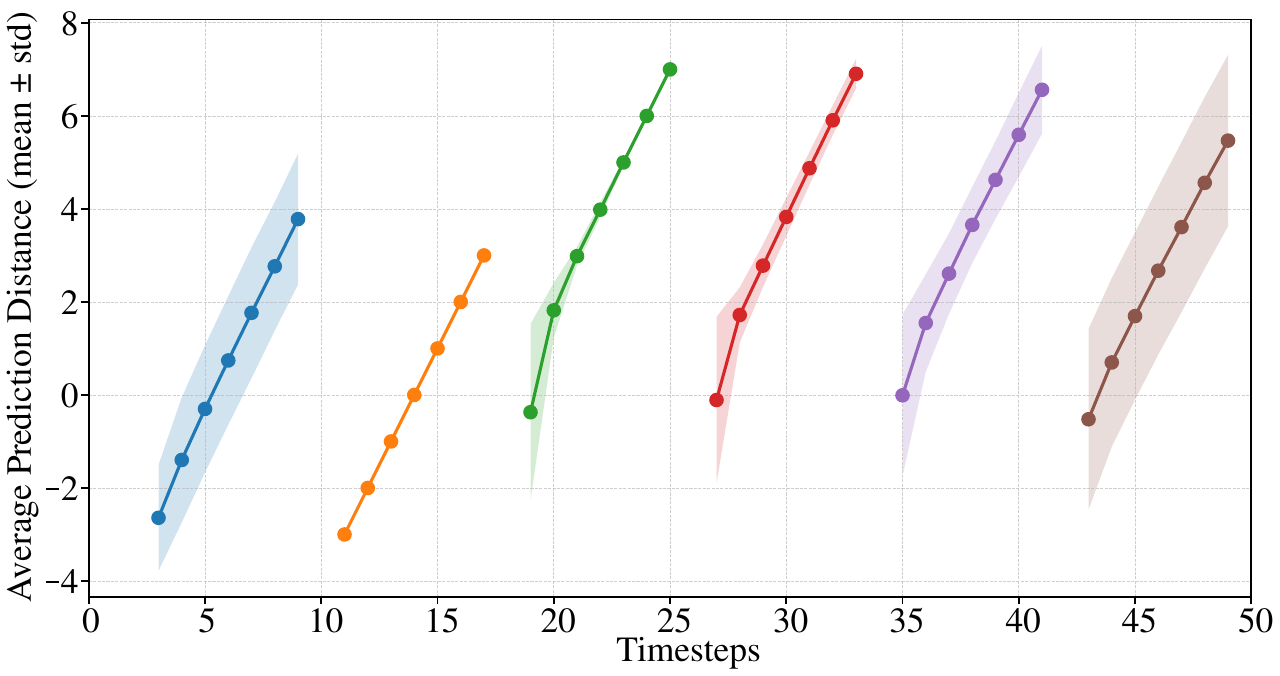}
        \caption{Distribution (mean and variance) of the selected prediction horizon over timesteps.}
        \label{fig:avg_distance_plot}
    \end{subfigure}
    \caption{Prediction patterns generated by TAP. Experiments were conducted on FLUX.1-dev with 50 sampling steps and an acceleration factor of \(N=8\).}
    \label{fig:prediction_patterns}
\end{figure*}

\paragraph{Token Prediction Patterns}
To further analyze the predictor assignments produced by TAP, we visualize TAP's averaged per-token predictions in Figure~\ref{fig:prediction_patterns}. We observe a clear temporal pattern. In the early stage (steps \(0\text{--}18\)), the model predominantly selects lower-order predictors. In the mid-to-late stage (steps \(18\text{--}50\)), the model shifts toward higher-order predictors and the variance of selected orders increases substantially, indicating a divergence in temporal-evolution patterns across tokens and motivating per-token assignment of different orders. A similar trend appears for prediction distance: its variance grows in the late stage (notably steps \(42\text{--}50\)). We also find that the model prefers smaller distances during the early stage (\(0\text{--}18\) steps) to remain closer to the convergence point. This behavior is expected because early-generation content is more uncertain, features evolve rapidly and non-smoothly, and Taylor-based forecasts computed from early steps suffer larger truncation errors and have a smaller radius of convergence. Together, these observations indicate that TAP dynamically adapts both predictor order and prediction horizon over time to better match changing generation dynamics.

\begin{figure*}[htbp]
\centering
\includegraphics[width=0.8\linewidth]{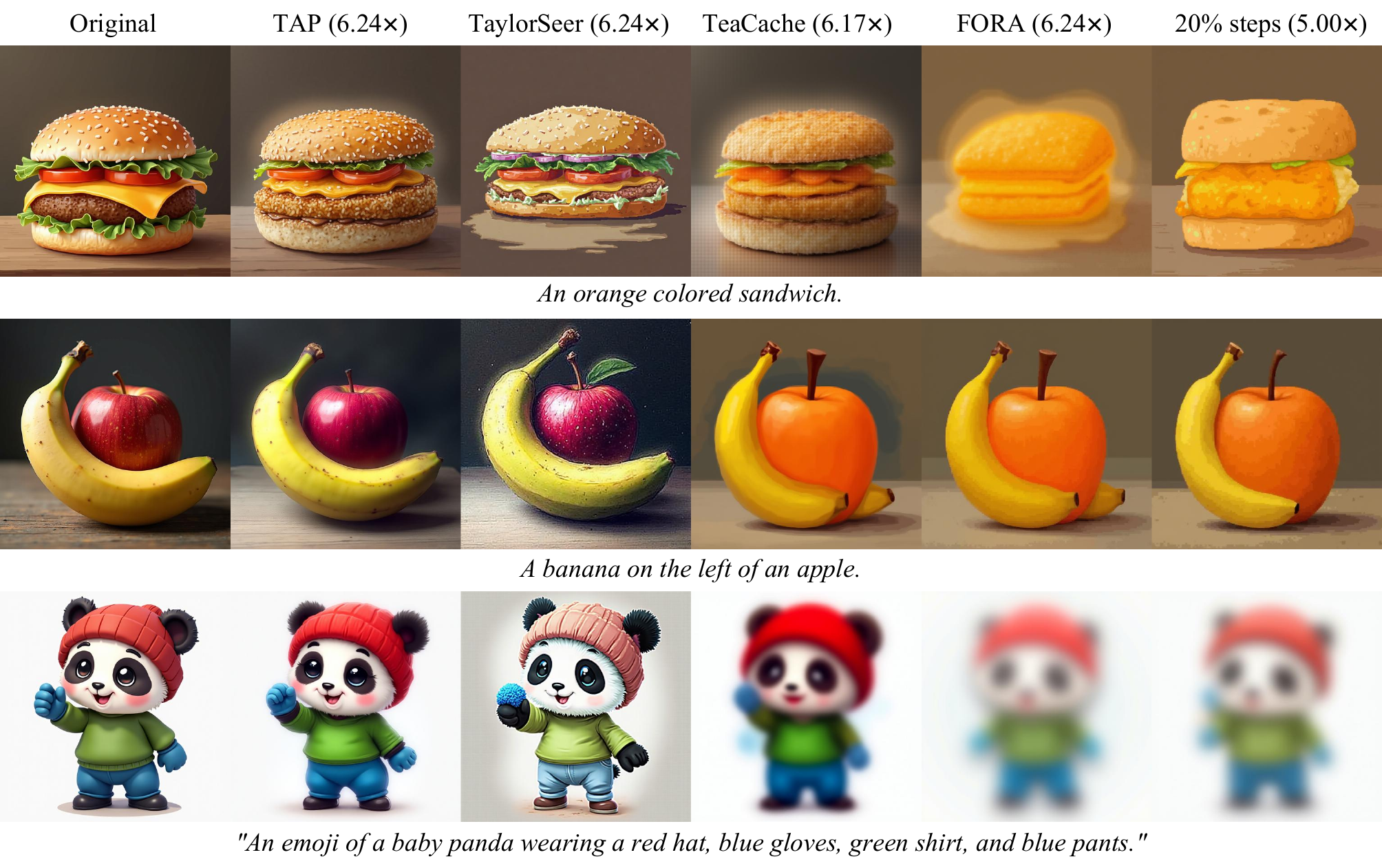}
\caption{Visualization results of FLUX.1-dev.}
\label{fig:more_vis_flux}
\vspace{-2mm}
\end{figure*}

\begin{figure*}[htbp]
\centering
\includegraphics[width=0.8\linewidth]{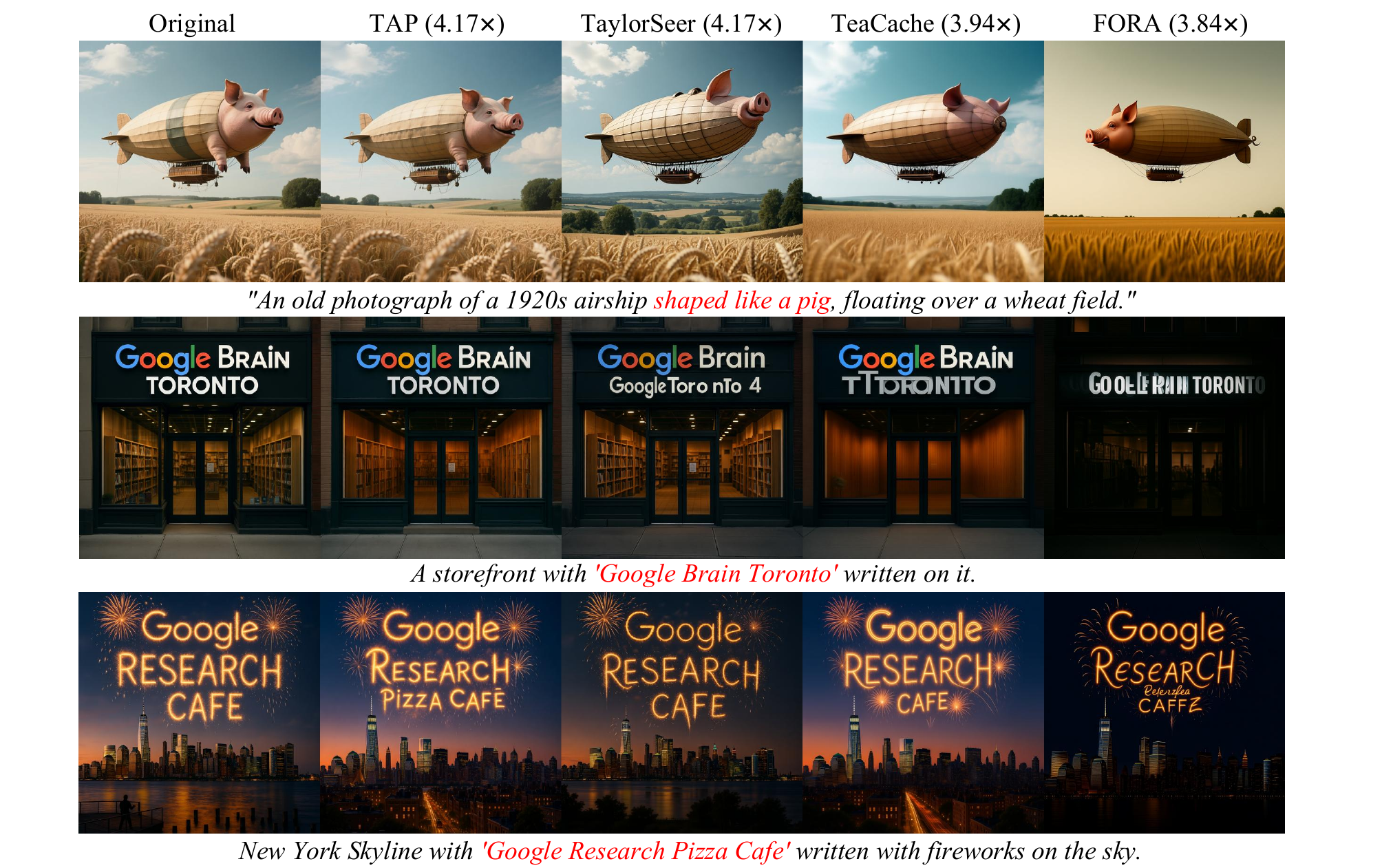}
\caption{Visualization results of Qwen-Image.}
\label{fig:more_vis_qwen}
\vspace{-2mm}
\end{figure*}

\begin{figure*}[htbp]
\centering
\includegraphics[width=0.8\linewidth]{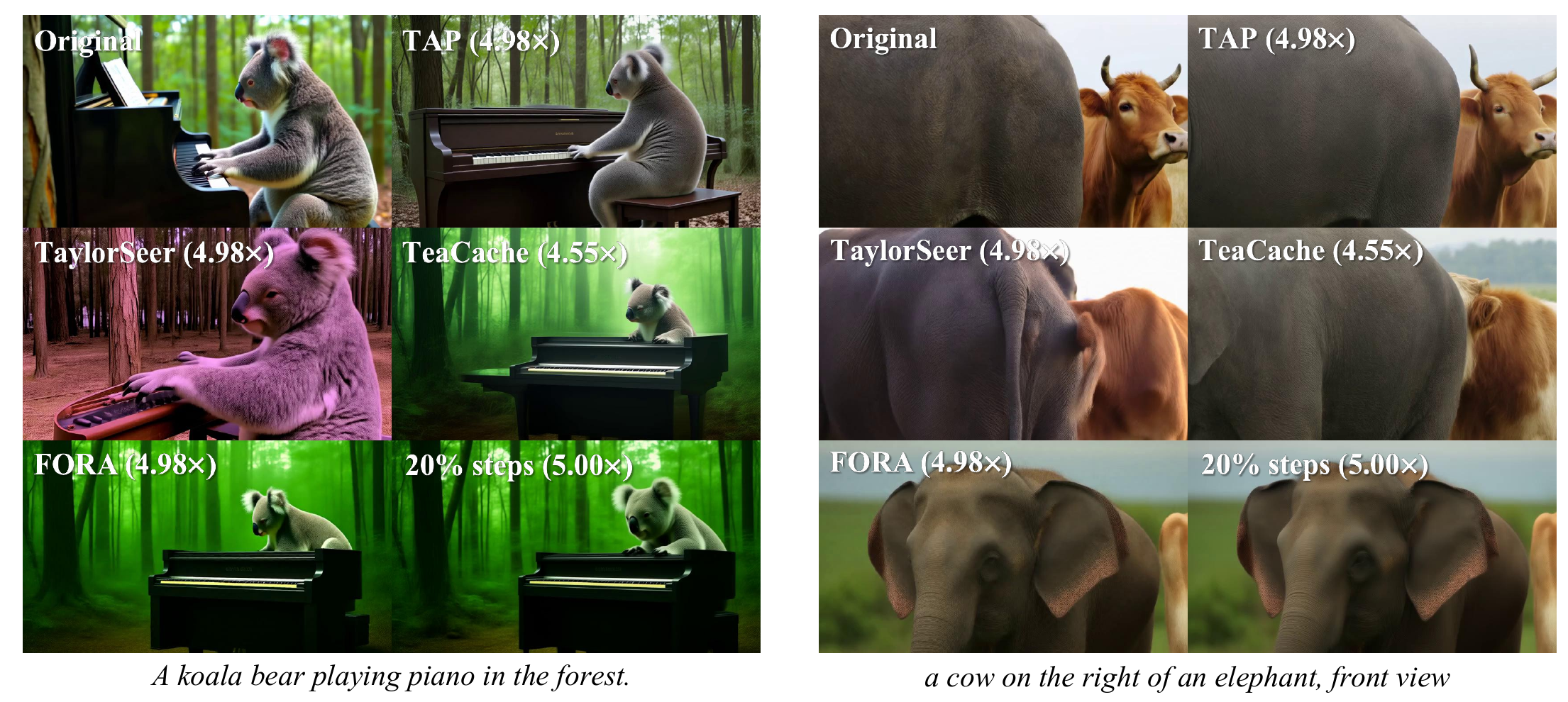}
\caption{Visualization results of HunyuanVideo.}
\label{fig:more_vis_hunyuan}
\vspace{-2mm}
\end{figure*}

\paragraph{Additional Qualitative Analysis}
We provide additional visualizations for the distilled variants in Figure~\ref{fig:vis_flux-schnell} to enable direct visual comparison.Besides, further visualizations are presented in Figures~\ref{fig:more_vis_flux}–\ref{fig:more_vis_hunyuan}. Compared with baseline methods, our approach produces results that are more faithful to the original model and better aligned with the text instructions. The baselines compared frequently exhibit visible distortions, loss of fine detail, or semantic misalignment.

%% file: main.bib
@string(ICCV= {Int. Conf. Comput. Vis.})

@string(ICCV  = {ICCV})

@article{structural_pruning_diffusion,
  title        = {Structural Pruning for Diffusion Models},
  author       = {Fang, Gongfan and Ma, Xinyin and Wang, Xinchao},
  year         = 2023,
  journal      = {arXiv preprint arXiv:2305.10924}
}

@article{salimans2022progressive,
  title        = {Progressive distillation for fast sampling of diffusion models},
  author       = {Salimans, Tim and Ho, Jonathan},
  year         = 2022,
  journal      = {arXiv preprint arXiv:2202.00512}
}

@inproceedings{songDDIM,
  title        = {Denoising Diffusion Implicit Models},
  author       = {Song, Jiaming and Meng, Chenlin and Ermon, Stefano},
  year         = 2021,
  booktitle    = {International Conference on Learning Representations}
}

@article{ho2020DDPM,
  title        = {Denoising diffusion probabilistic models},
  author       = {Ho, Jonathan and Jain, Ajay and Abbeel, Pieter},
  year         = 2020,
  journal      = {Advances in neural information processing systems},
  volume       = 33,
  pages        = {6840--6851}
}

@inproceedings{sohl2015deep,
  title        = {Deep unsupervised learning using nonequilibrium thermodynamics},
  author       = {Sohl-Dickstein, Jascha and Weiss, Eric and Maheswaranathan, Niru and Ganguli, Surya},
  year         = 2015,
  booktitle    = {International conference on machine learning},
  pages        = {2256--2265},
  organization = {PMLR}
}

@article{blattmann2023SVD,
  title        = {Stable video diffusion: Scaling latent video diffusion models to large datasets},
  author       = {Blattmann, Andreas and Dockhorn, Tim and Kulal, Sumith and Mendelevitch, Daniel and Kilian, Maciej and Lorenz, Dominik and Levi, Yam and English, Zion and Voleti, Vikram and Letts, Adam and others},
  year         = 2023,
  journal      = {arXiv preprint arXiv:2311.15127}
}

@article{zou2024accelerating,
  title        = {Accelerating Diffusion Transformers with Token-wise Feature Caching},
  author       = {Zou, Chang and Liu, Xuyang and Liu, Ting and Huang, Siteng and Zhang, Linfeng},
  year         = 2024,
  journal      = {arXiv preprint arXiv:2410.05317}
}

@article{lu2022dpm,
  title        = {Dpm-solver: A fast ode solver for diffusion probabilistic model sampling in around 10 steps},
  author       = {Lu, Cheng and Zhou, Yuhao and Bao, Fan and Chen, Jianfei and Li, Chongxuan and Zhu, Jun},
  year         = 2022,
  journal      = {Advances in Neural Information Processing Systems},
  volume       = 35,
  pages        = {5775--5787}
}

@inproceedings{
zheng2023dpmsolvervF,
title={{DPM}-Solver-v3: Improved Diffusion {ODE} Solver with Empirical Model Statistics},
author={Kaiwen Zheng and Cheng Lu and Jianfei Chen and Jun Zhu},
booktitle={Thirty-seventh Conference on Neural Information Processing Systems},
year={2023},
url={https://openreview.net/forum?id=9fWKExmKa0}
}

@article{li2024snapfusion,
  title        = {Snapfusion: Text-to-image diffusion model on mobile devices within two seconds},
  author       = {Li, Yanyu and Wang, Huan and Jin, Qing and Hu, Ju and Chemerys, Pavlo and Fu, Yun and Wang, Yanzhi and Tulyakov, Sergey and Ren, Jian},
  year         = 2024,
  journal      = {Advances in Neural Information Processing Systems},
  volume       = 36
}

@misc{flux2024,
  title        = {FLUX},
  author       = {Black Forest Labs},
  year         = 2024,
  howpublished = {\url{https://github.com/black-forest-labs/flux}}
}

@misc{liu2024timestep,
    title={Timestep Embedding Tells: It's Time to Cache for Video Diffusion Model},
    author={Feng Liu and Shiwei Zhang and Xiaofeng Wang and Yujie Wei and Haonan Qiu and Yuzhong Zhao and Yingya Zhang and Qixiang Ye and Fang Wan},
    year={2024},
    eprint={2411.19108},
    archivePrefix={arXiv},
    primaryClass={cs.CV}
}

@inproceedings{kim2025ditto,
  author = {Sungbin Kim and Hyunwuk Lee and Wonho Cho and Mincheol Park and Won Woo Ro},
  title = {Ditto: Accelerating Diffusion Model via Temporal Value Similarity},
  booktitle = {Proceedings of the 2025 IEEE International Symposium on High-Performance Computer Architecture (HPCA)},
  year = {2025},
  publisher = {IEEE},
}

@misc{DM,
	title = {Denoising {Diffusion} {Probabilistic} {Models}},
	url = {http://arxiv.org/abs/2006.11239},
	doi = {10.48550/arXiv.2006.11239},
	urldate = {2025-02-27},
	publisher = {arXiv},
	author = {Ho, Jonathan and Jain, Ajay and Abbeel, Pieter},
	month = dec,
	year = {2020},
	note = {arXiv:2006.11239 [cs]},
}

@online{liuReusingForecastingAccelerating2025,
  title = {From {{Reusing}} to {{Forecasting}}: {{Accelerating Diffusion Models}} with {{TaylorSeers}}},
  shorttitle = {From {{Reusing}} to {{Forecasting}}},
  author = {Liu, Jiacheng and Zou, Chang and Lyu, Yuanhuiyi and Chen, Junjie and Zhang, Linfeng},
  date = {2025-03-10},
  eprint = {2503.06923},
  eprinttype = {arXiv},
  eprintclass = {cs},
  doi = {10.48550/arXiv.2503.06923},
  url = {http://arxiv.org/abs/2503.06923},
  pubstate = {prepublished}
}

@article{liu2025freqca,
  title={Freqca: Accelerating diffusion models via frequency-aware caching},
  author={Liu, Jiacheng and Cai, Peiliang and Zhou, Qinming and Lin, Yuqi and Kong, Deyang and Huang, Benhao and Pan, Yupei and Xu, Haowen and Zou, Chang and Tang, Junshu and others},
  journal={arXiv preprint arXiv:2510.08669},
  year={2025}
}

@inproceedings{liu2025speca,
  title={Speca: Accelerating diffusion transformers with speculative feature caching},
  author={Liu, Jiacheng and Zou, Chang and Lyu, Yuanhuiyi and Ren, Fei and Wang, Shaobo and Li, Kaixin and Zhang, Linfeng},
  booktitle={Proceedings of the 33rd ACM International Conference on Multimedia},
  pages={10024--10033},
  year={2025}
}

@article{selvaraju2024fora,
  title        = {FORA: Fast-Forward Caching in Diffusion Transformer Acceleration},
  author       = {Selvaraju, Pratheba and Ding, Tianyu and Chen, Tianyi and Zharkov, Ilya and Liang, Luming},
  year         = 2024,
  journal      = {arXiv preprint arXiv:2407.01425}
}

@inproceedings{ma2024deepcache,
  title        = {Deepcache: Accelerating diffusion models for free},
  author       = {Ma, Xinyin and Fang, Gongfan and Wang, Xinchao},
  year         = 2024,
  booktitle    = {Proceedings of the IEEE/CVF Conference on Computer Vision and Pattern Recognition},
  pages        = {15762--15772}
}

@article{chen2024delta-dit,
  title        = {$\Delta$-DiT: A Training-Free Acceleration Method Tailored for Diffusion Transformers},
  author       = {Chen, Pengtao and Shen, Mingzhu and Ye, Peng and Cao, Jianjian and Tu, Chongjun and Bouganis, Christos-Savvas and Zhao, Yiren and Chen, Tao},
  year         = 2024,
  journal      = {arXiv preprint arXiv:2406.01125}
}

@inproceedings{bolya2023tomesd,
  title        = {Token merging for fast stable diffusion},
  author       = {Bolya, Daniel and Hoffman, Judy},
  year         = 2023,
  booktitle    = {Proceedings of the IEEE/CVF conference on computer vision and pattern recognition},
  pages        = {4599--4603}
}

@article{lu2022dpm++,
  title        = {Dpm-solver++: Fast solver for guided sampling of diffusion probabilistic models},
  author       = {Lu, Cheng and Zhou, Yuhao and Bao, Fan and Chen, Jianfei and Li, Chongxuan and Zhu, Jun},
  year         = 2022,
  journal      = {arXiv preprint arXiv:2211.01095}
}

@inproceedings{kim2024tofu,
  title        = {Token fusion: Bridging the gap between token pruning and token merging},
  author       = {Kim, Minchul and Gao, Shangqian and Hsu, Yen-Chang and Shen, Yilin and Jin, Hongxia},
  year         = 2024,
  booktitle    = {Proceedings of the IEEE/CVF Winter Conference on Applications of Computer Vision},
  pages        = {1383--1392}
}

@inproceedings{shang2023post,
  title        = {Post-training quantization on diffusion models},
  author       = {Shang, Yuzhang and Yuan, Zhihang and Xie, Bin and Wu, Bingzhe and Yan, Yan},
  year         = 2023,
  booktitle    = {Proceedings of the IEEE/CVF conference on computer vision and pattern recognition},
  pages        = {1972--1981}
}

@inproceedings{huang2024vbench,
  title        = {Vbench: Comprehensive benchmark suite for video generative models},
  author       = {Huang, Ziqi and He, Yinan and Yu, Jiashuo and Zhang, Fan and Si, Chenyang and Jiang, Yuming and Zhang, Yuanhan and Wu, Tianxing and Jin, Qingyang and Chanpaisit, Nattapol and others},
  year         = 2024,
  booktitle    = {Proceedings of the IEEE/CVF Conference on Computer Vision and Pattern Recognition},
  pages        = {21807--21818}
}

@article{hessel2021clipscore,
  title        = {Clipscore: A reference-free evaluation metric for image captioning},
  author       = {Hessel, Jack and Holtzman, Ari and Forbes, Maxwell and Bras, Ronan Le and Choi, Yejin},
  year         = 2021,
  journal      = {arXiv preprint arXiv:2104.08718}
}

@inproceedings{
meng2022on,
title={On Distillation of Guided Diffusion Models},
author={Chenlin Meng and Ruiqi Gao and Diederik P Kingma and Stefano Ermon and Jonathan Ho and Tim Salimans},
booktitle={NeurIPS 2022 Workshop on Score-Based Methods},
year={2022},
url={https://openreview.net/forum?id=6QHpSQt6VR-}
}

@INPROCEEDINGS{10377259,
  author={Li, Xiuyu and Liu, Yijiang and Lian, Long and Yang, Huanrui and Dong, Zhen and Kang, Daniel and Zhang, Shanghang and Keutzer, Kurt},
  booktitle={2023 IEEE/CVF International Conference on Computer Vision (ICCV)}, 
  title={Q-Diffusion: Quantizing Diffusion Models}, 
  year={2023},
  volume={},
  number={},
  pages={17489-17499},
  doi={10.1109/ICCV51070.2023.01608}}

@inproceedings{
yuan2024ditfastattn,
title={Di{TF}astAttn: Attention Compression for Diffusion Transformer Models},
author={Zhihang Yuan and Hanling Zhang and Lu Pu and Xuefei Ning and Linfeng Zhang and Tianchen Zhao and Shengen Yan and Guohao Dai and Yu Wang},
booktitle={The Thirty-eighth Annual Conference on Neural Information Processing Systems},
year={2024},
url={https://openreview.net/forum?id=51HQpkQy3t}
}

@misc{zhu2024dipgo,
    title={DiP-GO: A Diffusion Pruner via Few-step Gradient Optimization},
    author={Haowei Zhu and Dehua Tang and Ji Liu and Mingjie Lu and Jintu Zheng and Jinzhang Peng and Dong Li and Yu Wang and Fan Jiang and Lu Tian and Spandan Tiwari and Ashish Sirasao and Jun-Hai Yong and Bin Wang and Emad Barsoum},
    year={2024},
    eprint={2410.16942},
    archivePrefix={arXiv},
    primaryClass={cs.CV}
}

@misc{wu2025qwenimagetechnicalreport,
      title={Qwen-Image Technical Report}, 
      author={Chenfei Wu and Jiahao Li and Jingren Zhou and Junyang Lin and Kaiyuan Gao and Kun Yan and Sheng-ming Yin and Shuai Bai and Xiao Xu and Yilei Chen and Yuxiang Chen and Zecheng Tang and Zekai Zhang and Zhengyi Wang and An Yang and Bowen Yu and Chen Cheng and Dayiheng Liu and Deqing Li and Hang Zhang and Hao Meng and Hu Wei and Jingyuan Ni and Kai Chen and Kuan Cao and Liang Peng and Lin Qu and Minggang Wu and Peng Wang and Shuting Yu and Tingkun Wen and Wensen Feng and Xiaoxiao Xu and Yi Wang and Yichang Zhang and Yongqiang Zhu and Yujia Wu and Yuxuan Cai and Zenan Liu},
      year={2025},
      eprint={2508.02324},
      archivePrefix={arXiv},
      primaryClass={cs.CV},
      url={https://arxiv.org/abs/2508.02324}, 
}

@misc{StableDiffusion,
	title = {High-{Resolution} {Image} {Synthesis} with {Latent} {Diffusion} {Models}},
	url = {http://arxiv.org/abs/2112.10752},
	doi = {10.48550/arXiv.2112.10752},

	urldate = {2025-02-27},
	publisher = {arXiv},
	author = {Rombach, Robin and Blattmann, Andreas and Lorenz, Dominik and Esser, Patrick and Ommer, Björn},
	month = apr,
	year = {2022},
	note = {arXiv:2112.10752 [cs]},
}

@misc{DiT,
	title = {Scalable {Diffusion} {Models} with {Transformers}},
	url = {http://arxiv.org/abs/2212.09748},
	doi = {10.48550/arXiv.2212.09748},
	urldate = {2025-02-27},
	publisher = {arXiv},
	author = {Peebles, William and Xie, Saining},
	month = mar,
	year = {2023},
	note = {arXiv:2212.09748 [cs]},

}

@misc{zhangUnreasonableEffectivenessDeep2018,
  title = {The {{Unreasonable Effectiveness}} of {{Deep Features}} as a {{Perceptual Metric}}},
  author = {Zhang, Richard and Isola, Phillip and Efros, Alexei A. and Shechtman, Eli and Wang, Oliver},
  year = {2018},
  number = {arXiv:1801.03924},
  eprint = {1801.03924},
  primaryclass = {cs},
  publisher = {arXiv},
  doi = {10.48550/arXiv.1801.03924},
  archiveprefix = {arXiv}
}

@misc{liuTaylorSeer2025,
  title = {From Reusing to Forecasting: Accelerating Diffusion Models with TaylorSeers},
  shorttitle = {TaylorSeer},
  author = {Liu, Jiacheng and Zou, Chang and Lyu, Yuanhuiyi and Chen, Junjie and Zhang, Linfeng},
  year = {2025},
  number = {arXiv:2503.06923},
  eprint = {2503.06923},
  primaryclass = {cs.CV},
  publisher = {arXiv},
  doi = {10.48550/arXiv.2503.06923},
  archiveprefix = {arXiv}
}

@misc{zhengFoCa2025,
  title = {Forecast then Calibrate: Feature Caching as ODE for Efficient Diffusion Transformers},
  shorttitle = {FoCa},
  author = {Zheng, Shikang and Feng, Liang and Wang, Xinyu and Zhou, Qinming and Cai, Peiliang and Zou, Chang and Liu, Jiacheng and Lin, Yuqi and Chen, Junjie and Ma, Yue and Zhang, Linfeng},
  year = {2025},
  number = {arXiv:2508.16211},
  eprint = {2508.16211},
  primaryclass = {cs.CV},
  publisher = {arXiv},
  doi = {10.48550/arXiv.2508.16211},
  archiveprefix = {arXiv}
}

@misc{fengHiCache2025,
  title = {HiCache: Training-free Acceleration of Diffusion Models via Hermite Polynomial-based Feature Caching},
  shorttitle = {HiCache},
  author = {Feng, Liang and Zheng, Shikang and Liu, Jiacheng and Lin, Yuqi and Zhou, Qinming and Cai, Peiliang and Wang, Xinyu and Chen, Junjie and Zou, Chang and Ma, Yue and Zhang, Linfeng},
  year = {2025},
  number = {arXiv:2508.16984},
  eprint = {2508.16984},
  primaryclass = {cs.CV},
  publisher = {arXiv},
  doi = {10.48550/arXiv.2508.16984},
  archiveprefix = {arXiv}
}

@inproceedings{saharia2022drawbench,
  title        = {Photorealistic Text-to-Image Diffusion Models with Deep Language Understanding},
  author       = {Saharia, Chitwan and Chan, William and Saxena, Saurabh and Li, Lala and Whang, Jay and Denton, Emily and Ghasemipour, Seyed Kamyar Seyed and Ayan, Burcu Karagol and Mahdavi, S. Sara and Lopes, Rapha Gontijo and Salimans, Tim and Ho, Jonathan and Fleet, David J. and Norouzi, Mohammad},
  booktitle    = {Proceedings of the IEEE/CVF Conference on Computer Vision and Pattern Recognition},
  pages        = {24219--24238},
  year         = 2022
}

@misc{xu2023imagereward,
  title        = {ImageReward: Learning and Evaluating Human Preferences for Text-to-Image Generation},
  author       = {Xu, Jiazheng and Liu, Xiao and Wu, Yuchen and Tong, Yuxuan and Li, Qinkai and Ding, Ming and Tang, Jie and Dong, Yuxiao},
  year         = 2023,
  eprint       = {2304.05977},
  archivePrefix = {arXiv},
  primaryClass = {cs.CV},
  url          = {https://arxiv.org/abs/2304.05977}
}

@article{wang2004imagequality,
  title = {Image Quality Assessment: From Error Visibility to Structural Similarity},
  author = {Wang, Zhou and Bovik, A.C. and Sheikh, H.R. and Simoncelli, E.P.},
  year = {2004},
  journal = {IEEE Transactions on Image Processing},
  volume = {13},
  number = {4},
  pages = {600--612},
  doi = {10.1109/TIP.2003.819861}
}

@article{sagi2018ensemble,
  title={Ensemble learning: A survey},
  author={Sagi, Omer and Rokach, Lior},
  journal={Wiley interdisciplinary reviews: data mining and knowledge discovery},
  volume={8},
  number={4},
  pages={e1249},
  year={2018},
  publisher={Wiley Online Library}
}

@inproceedings{yin2024improved,
    title={Improved Distribution Matching Distillation for Fast Image Synthesis},
    author={Yin, Tianwei and Gharbi, Micha{\"e}l and Park, Taesung and Zhang, Richard and Shechtman, Eli and Durand, Fredo and Freeman, William T},
    booktitle={NeurIPS},
    year={2024}
}

@misc{cheng2025paraattention,
  author = {Cheng, Zeyi},
  title = {ParaAttention: Context Parallel Attention that Accelerates Dit Model Inference with Dynamic Caching},
  howpublished = {\url{https://github.com/chengzeyi/ParaAttention}},
  year = {2025},
  note = {Accessed: YYYY-MM-DD}
}

@inproceedings{levin2025differential,
  title={Differential diffusion: Giving each pixel its strength},
  author={Levin, Eran and Fried, Ohad},
  booktitle={Computer Graphics Forum},
  pages={e70040},
  year={2025},
  organization={Wiley Online Library}
}

@article{kong2024hunyuanvideo,
  title={Hunyuanvideo: A systematic framework for large video generative models},
  author={Kong, Weijie and Tian, Qi and Zhang, Zijian and Min, Rox and Dai, Zuozhuo and Zhou, Jin and Xiong, Jiangfeng and Li, Xin and Wu, Bo and Zhang, Jianwei and others},
  journal={arXiv preprint arXiv:2412.03603},
  year={2024}
}

@article{bu2025dicache,
  title={DiCache: Let Diffusion Model Determine Its Own Cache},
  author={Bu, Jiazi and Ling, Pengyang and Zhou, Yujie and Wang, Yibin and Zang, Yuhang and Wu, Tong and Lin, Dahua and Wang, Jiaqi},
  journal={arXiv preprint arXiv:2508.17356},
  year={2025}
}

@inproceedings{yu2025ab,
  title={Ab-cache: Training-free acceleration of diffusion models via adams-bashforth cached feature reuse},
  author={Yu, Zichao and Zou, Zhen and Shao, Guojiang and Zhang, Chenwei and Xu, Shengze and Huang, Jie and Zhao, Feng and Cun, Xiaodong and Zhang, Wenyi},
  booktitle={Proceedings of the 33rd ACM International Conference on Multimedia},
  pages={10408--10417},
  year={2025}
}

@online{calflops,
  author = {xiaoju ye},
  title = {calflops: a FLOPs and Params calculate tool for neural networks in pytorch framework},
  year = 2023,
  url = {https://github.com/MrYxJ/calculate-flops.pytorch},
}
